\documentclass[runningheads]{llncs}
\usepackage{graphicx}

\usepackage{tikz}
\usepackage{comment}
\usepackage{amsmath,amssymb} 
\usepackage{color}

\usepackage{graphicx}
\usepackage{booktabs}
\usepackage{epsfig}
\usepackage{bm}
\usepackage{caption} 
\usepackage[labelfont=bf, font=footnotesize]{caption}
\usepackage{color}
\usepackage{multirow}
\usepackage[ruled,vlined]{algorithm2e}
\usepackage{threeparttable}
\usepackage{url}
\usepackage{colortbl}
\usepackage{wrapfig}
\usepackage[lofdepth,lotdepth]{subfig}
\usepackage{cite}

\usepackage[accsupp]{axessibility}  

\newcommand{\myparagraph}[1]{\noindent\textbf{#1}}

\def\ie{\textit{i.e. }}
\def\eg{\textit{e.g. }}

\definecolor{Gray}{gray}{0.9}
\definecolor{green}{rgb}{0.55, 0.71, 0.0}
\definecolor{amaranth}{rgb}{0.9, 0.17, 0.31}
\definecolor{amber}{rgb}{1.0, 0.49, 0.0}
\definecolor{azure}{rgb}{0.0, 0.5, 1.0}
\definecolor{byzantine}{rgb}{0.74, 0.2, 0.64}
\definecolor{forestGreen}{rgb}{0.0, 0.54, 0.0}
\definecolor{blue}{rgb}{0.43, 0.71, 0.88}
\definecolor{pink}{rgb}{0.85, 0.44, 0.58}

\newcommand{\name}{EgoBody}
\newcommand{\interactee}{interactee}

\usepackage[pagebackref=true,breaklinks=true,letterpaper=true,colorlinks,bookmarks=false]{hyperref}

\begin{document}
\pagestyle{headings}
\mainmatter
\def\ECCVSubNumber{4963}  

\title{EgoBody: Human Body Shape and Motion of Interacting People from Head-Mounted Devices} 

\titlerunning{EgoBody: Human Body Shape and Motion of Interacting People from HMD}
%


\authorrunning{S. Zhang et al.}
%

\newcommand*{\affaddr}[1]{#1} 
\newcommand*{\affmark}[1][*]{\textsuperscript{#1}}
\author{
Siwei Zhang\affmark[1] \quad
Qianli Ma\affmark[1]\quad
Yan Zhang\affmark[1] \quad
Zhiyin Qian\affmark[1]  \quad 
Taein Kwon\affmark[1]  \quad 
Marc Pollefeys\affmark[1,2]  \quad
Federica Bogo\affmark[2]\thanks{Now at Meta Reality Labs Research.}  \quad
Siyu Tang\affmark[1]}
\institute{
\affaddr{\affmark[1]ETH Z\"urich} \quad 
\affaddr{\affmark[2]Microsoft} \quad \\
\email{\{siwei.zhang, qianli.ma, yan.zhang, taein.kwon, marc.pollefeys, siyu.tang\}@inf.ethz.ch} \quad 
\email{zhqian@ethz.ch} \quad 
\email{fbogo@fb.com}
}

\maketitle

\begin{abstract}
Understanding social interactions
from \textit{egocentric} views is crucial for many applications, ranging from assistive robotics to AR/VR. Key to reasoning about interactions is 
to understand the body pose and motion of the interaction partner from the egocentric view.
However, research in this area is severely hindered by the lack of datasets.
Existing datasets are limited in terms of either size, capture/annotation modalities, ground-truth quality, or interaction diversity.
We fill this gap by proposing \name{}, a novel 
large-scale dataset for human pose, shape and motion estimation from egocentric views, during interactions in complex 3D scenes. 
We employ Microsoft HoloLens2 headsets to record rich egocentric data streams (including RGB, depth, eye gaze, head and hand tracking). To obtain accurate 3D ground truth, we calibrate the headset with a multi-Kinect rig and fit expressive SMPL-X body meshes to multi-view RGB-D frames, reconstructing 3D human shapes and poses relative to the scene, over time. 
We collect 125 sequences, spanning diverse interaction scenarios, and propose the first benchmark for 3D full-body pose and shape estimation of the interaction partner from egocentric views. 
We extensively evaluate state-of-the-art methods, highlight their limitations in the egocentric scenario, and address such limitations leveraging our high-quality annotations.
Data and code are available at \url{https://sanweiliti.github.io/egobody/egobody.html}.

\textbf{Keywords:} pose estimation, egocentric view, motion capture, dataset
\end{abstract}


\section{Introduction}
\label{sec:intro}
\begin{figure}[t]
    \centering
    \includegraphics[width=\linewidth]{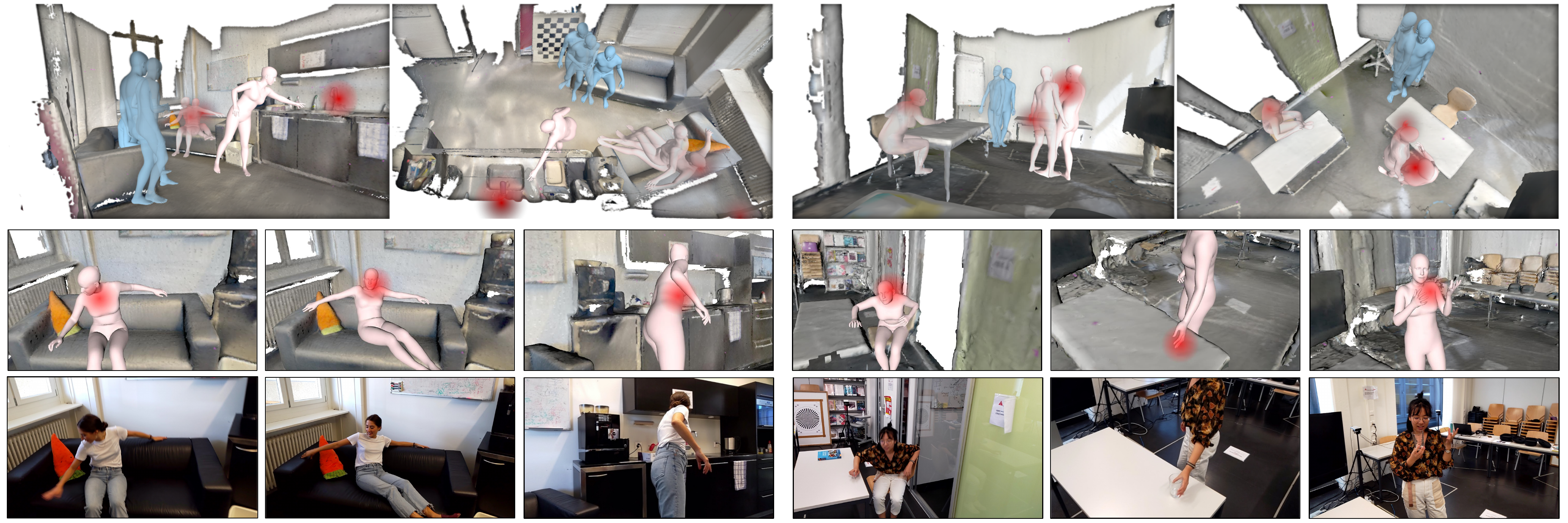}
    \caption{\footnotesize \name{} is a large-scale dataset capturing ground-truth 3D human motions during social interactions in 3D scenes.
  Given two interacting subjects, we leverage a lightweight multi-camera rig to reconstruct their 3D shape and pose over time (top row). One of the subjects (\textcolor{blue}{blue}) wears a head-mounted device, synchronized with the rig, capturing egocentric multi-modal data like eye gaze tracking (\textcolor{red}{red} circles in first two rows) and RGB images (bottom).}
    \label{fig:teaser}
\end{figure}

Humans constantly interact and communicate with each other; understanding our social interaction partners' motions, intentions and emotions is almost instinctive for us. 
However, the same does not hold for machines. 
A first step towards automated human interaction understanding is the estimation of the 3D body pose, shape and motion of the social interaction partner (``\textit{\interactee{}}'') from egocentric views, \eg from head-mounted devices (HMD). 
Addressing this challenging problem is crucial for many applications, ranging from assistive robotics to Augmented and Virtual Reality (AR/VR), where sensors typically perceive the \interactee{} from the egocentric view. 
Despite its importance, the problem has received little attention in the literature so far. 
While there are a large number of methods for full-body pose (and sometimes also shape) estimation from RGB(D) frames~\cite{choi2021beyond,fang2021reconstructing,hassan2019resolving,kanazawa2018end,kanazawa2019learning,kocabas2020vibe,kolotouros2019learning,luo20203d,pavlakos2019expressive,sun2019human,weng2021holistic,xu2019denserac,yuan2021simpoe,zanfir2020weakly,zhang2021body,Zhang:ICCV:2021}, they tend to perform poorly on data captured with an HMD (see Sec.~\ref{sec:experiment}). Indeed, this setup brings its own unique challenges, which most methods have not explicitly addressed so far.
Any method aiming at understanding the pose and shape of the \interactee{} must deal with severe body truncations, motion blur (exacerbated by the embodied movement of the HMD), people entering/exiting the field of view, to name a few.

\begin{table*}[t!]
\centering
\caption{Comparison with existing image-based datasets with 3D human pose annotations.
``Fr.\#'' denotes frame numbers.
``3rd-PV'' and ``Ego'' refers to the third-person-view and egocentric view, respectively.
``Mesh'' refers to the body mesh.
``Interact'' refers to social interactions. ``Global-Cfg.'' refers to global translation and rotation. 
}
\begin{threeparttable}
\begin{tabular}{crrccccccc}
  \toprule[1pt]

  Dataset & Fr.\# & Sub.\# & 3rd-PV & Ego & Mesh & Gaze & 3D-Scene & Interact & Global-Cfg.\\
 
  \midrule
  TNT15~\cite{vonPon2016a} & 13k & 4 & \checkmark & & \checkmark & & & & \checkmark \\
  \rowcolor{Gray}
  3DPW~\cite{von2018recovering}  & 51k & 7 & \checkmark &  & \checkmark\tnote{*} & & & \checkmark & \\
  PROX~\cite{hassan2019resolving} & 100k & 20 & \checkmark &  & \checkmark\tnote{*} & & \checkmark & &\checkmark \\
  \rowcolor{Gray}
  Panoptic~\cite{joo2017panoptic}  & 297k & 180+ & \checkmark &  & & &  & \checkmark & \checkmark \\
  HUMBI~\cite{yu2020humbi}  & 380k & 772 & \checkmark & & \checkmark\tnote{*} & \checkmark & & & \checkmark \\
    \rowcolor{Gray}
  TotalCapture~\cite{Trumble:BMVC:2017} & 1,900k  & 5 & \checkmark & & & & & & \checkmark \\
  Human3.6M~\cite{ionescu2013human3} & 3,600k & 11 & \checkmark & & \checkmark & & & & \checkmark \\

  \rowcolor{Gray}
  Mo2Cap2~\cite{xu2019mo}  & 15k & 5 & & \checkmark & & & & & \\
  You2Me~\cite{ng2020you2me}  & 150k & 10 &  &\checkmark  & & & & \checkmark & \\
    \rowcolor{Gray}
  HPS~\cite{guzov2021human}  & 300k & 7 &  & \checkmark & \checkmark\tnote{*} & & \checkmark & &\checkmark \\
  \midrule
  Ours & 220k & 36 &  \checkmark & \checkmark  & \checkmark\tnote{*} & \checkmark & \checkmark & \checkmark & \checkmark \\

  \bottomrule[1pt]
 \end{tabular}
 \begin{tablenotes}
        \item[*] Body Mesh defined by parametric body models.
      \end{tablenotes}
 \end{threeparttable}
\label{tab:existing-datasets}
\end{table*}

A reason for such limited attention is the lack of data. 
On one hand, most human motion datasets are captured by \textit{third-person-view} cameras without egocentric frames~\cite{fieraru2020three,hassan2019resolving,ionescu2013human3,joo2019towards,joo2017panoptic,mehta2017monocular,Trumble:BMVC:2017,zhang20204d}, which do not faithfully replicate AR/VR scenarios; most capture only one subject at a time, without interactions~\cite{hassan2019resolving,ionescu2013human3}. 
On the other hand, existing \textit{egocentric} datasets are limited in terms of annotation modalities, scale and interaction diversity.
They either focus on coarse-level interaction/action labels~\cite{fathi2012social,li2018eye,narayan2014action,pirsiavash2012detecting,sigurdsson2018actor}, or provide only the camera wearer's pose without considering~\cite{tome2020selfpose,tome2019xr,xu2019mo,yuan2019ego}, or with very limited data involving~\cite{guzov2021human}, the \interactee{}.
You2Me~\cite{ng2020you2me} collects egocentric RGB frames of two-people interactions, annotated with 3D skeletons, without 3D scene context, nor the body shape.
Recently, Ego4D~\cite{ego4d} collects a large amount of egocentric videos for various tasks including action and social interaction understanding, but without 3D ground truth for human pose, shape and motions.

To fill this gap, we propose \name{}, a unique, large-scale egocentric dataset capturing high-quality 3D human motions during social interactions. 
We focus on 2-people interaction cases, 
and define interaction scenarios based on the social interaction categories studied in sociology~\cite{nisbet1970social}. 
Unlike most existing datasets that only provide RGB streams, 
\name{} collects egocentric multi-modal data, with accurate 3D human shape, pose and motion ground-truth for both interacting subjects, accompanied by eye gaze tracking for the camera wearer.
Furthermore, \name{} includes accurate 3D scene reconstructions, providing a holistic and consistent 3D understanding of the physical world around the camera wearer. 

The egocentric data is captured with a Microsoft HoloLens2 headset~\cite{hololens2}, which provides rich multi-modal streams: RGB, depth, head, hand and eye gaze tracking, correlated in space and time.
In particular, eye gaze carries vital information about human attention during interactions.
By providing eye gaze tracking synchronized with other modalities, EgoBody opens the door to study relationships between human attention, interactions and motions. 
We obtain high-quality 3D human shape and motion annotations in an automated way, by leveraging a marker-less motion capture approach. Namely, we utilize a multi-camera rig consisting of multiple Azure Kinects~\cite{azurekinect} as our motion capture system.

However, combining raw data streams from the egocentric- and the third-person-view remains highly challenging due to hardware limitations. 
Specifically, the Kinect-HoloLens2 calibration exhibit inaccuracies due to not perfectly accurate factory calibration and tracking drift. 
We address this by proposing a refinement scheme based on body keypoints. 
With carefully calibrated data, we further build an efficient motion capture pipeline based on \cite{Zhang:ICCV:2021} to fit the SMPL-X body model~\cite{pavlakos2019expressive} to multi-view and egocentric RGB-D data, reconstructing accurate 3D full-body meshes for both the camera wearer and the \interactee{}. 
In this way, we get accurate and well calibrated ground truth across all sensor coordinates, as well as the world coordinate, which is not available in most existing datasets.
The setup is lightweight and easy to deploy in various environments.

With~\name{} we propose the first benchmark for 3D human pose and shape estimation (3DHPS) of the \interactee{}, in interactions captured by the HMD. 
By evaluating state-of-the-art 3DHPS methods on the \name{}'s test set, we carefully analyze and highlight the limitations of existing methods in this egocentric setup. 
We show the usefulness of \name{} by fine-tuning three recent methods~\cite{joo2020eft,kolotouros2019learning,lin2021end-to-end} on its training set, obtaining significantly improved performance on our \textit{test set}.
Finally, in a cross-dataset evaluation we show how models fine-tuned on \name{} also achieve a better performance on the \textit{You2Me} dataset~\cite{ng2020you2me}.

\myparagraph{Contributions.} In summary, we:
\textbf{(1)} provide the first large-scale egocentric dataset, EgoBody, comprising both egocentric- and third-person-view multi-modal data, annotated with high-quality 3D ground-truth motions for \textit{both} interacting people and 3D scene reconstructions; 
\textbf{(2)} extensively evaluate state-of-the-art 3DHPS methods on our test set, showing their shortcomings in this egocentric setup and providing insights for future methods in this direction;
\textbf{(3)} show the usefulness of our training set: a simple fine-tuning on it significantly improves existing methods' performance and robustness on both our test set \emph{and a different egocentric dataset};
\textbf{(4)} provide the first benchmark for 3DHPS estimation of the \interactee{} in the egocentric view during social interactions.

\section{Related Work}
\label{sec:related}

\myparagraph{Datasets for 3D human pose, motion and interactions.} A large number of datasets focus on 3D human pose and motion from \emph{third-person-views}~\cite{fieraru2020three,hassan2019resolving,ionescu2013human3,joo2019towards,joo2017panoptic,cmu,von2018recovering,vonPon2016a,mehta2017monocular,Patel:CVPR:2021,Trumble:BMVC:2017,yu2020humbi,zhang20204d}. 
For example, Human3.6M~\cite{ionescu2013human3} and AMASS~\cite{mahmood2019amass} use optical marker-based motion capture to collect large amounts of high-quality 3D motion sequences; they are limited to constrained studio setups and images -- when available -- are polluted by markers. PROX~\cite{hassan2019resolving} performs marker-less capture of people moving in 3D scenes from monocular RGB-D, without human-human interactions.
The quality of the reconstructed motion is further improved by LEMO~\cite{Zhang:ICCV:2021}.
The Panoptic Studio datasets~\cite{joo2019towards,joo2017panoptic,joo2018total,xiang2019monocular} capture interactions between people using a multi-view camera system, annotated with body and hand 3D joints plus facial landmarks.
CHI3D~\cite{fieraru2020three} focuses on close human-human contacts, using a motion capture system to extract ground-truth 3D skeletons. 3DPW~\cite{von2018recovering} reconstructs the 3D shape and motion of people by fitting SMPL~\cite{loper2015smpl} to IMU data and RGB images captured with a hand-held camera, without 3D environment reconstruction. 
None of these datasets provides egocentric data. 

Among datasets for \emph{egocentric} vision, a lot of attention has been put on hand-object interactions and action recognition, often without 3D ground-truth~\cite{damen2018scaling,kay2017kinetics,kwon2021h2o,li2018eye,narayan2014action,pirsiavash2012detecting,ryoo2013first,sigurdsson2018actor,yonetani2016recognizing, ogaki2012coupling, kitani2011fast, bambach2015lending, zhang2022can, fathi2011understanding, kazakos2019epic, damen2022rescaling}. Mo2Cap2~\cite{xu2019mo} and xR-EgoPose~\cite{tome2019xr,tome2020selfpose} provide image-3D skeleton pairs for egocentric body pose prediction of the camera wearer, without the \interactee{} involved. 
HPS~\cite{guzov2021human} reconstructs the body pose and shape of the camera wearer moving in large 3D scenes; only a few frames include interactions with an \interactee{}.
You2Me~\cite{ng2020you2me} provides 3D skeletons for both interacting people paired with images captured with a chest-mounted camera plus external cameras; there are no body shape or 3D scene annotations.
EgoMoCap~\cite{liu20204d} analyzes the \interactee{} body shape and pose in outdoor social scenarios capturing only the egocentric RGB stream.

Table~\ref{tab:existing-datasets} compares \name{} with the most related human motion datasets. \name{} is the first motion capture dataset that collects calibrated egocentric- and third-person-view images, with various interaction scenarios, multi-modal data and rich 3D ground-truth. 
Additionally, \name{} provides the camera wearer's eye gaze to facilitate potential social interaction studies which jointly analyze human attention and motion.

\myparagraph{3D human pose estimation.}
The problem of estimating 3D human pose from \emph{third-person-view} 
RGB(D) images has been extensively studied in the literature -- either from single frames~\cite{grauman2003inferring,agarwal2005recovering,bualan2008naked,tan2017indirect,tung2017self,kanazawa2018end,hassan2019resolving,kolotouros2019learning,kolotouros2019cmr,pavlakos2019expressive,guler2019holopose,omran2018neural,xu2019denserac,fang2021reconstructing,zhang2021body,weng2021holistic,zhou2021monocular,li2021hybrik,wandt2021canonpose,lin2021end-to-end,Moon_2020_ECCV_I2L-MeshNet,Choi_2020_ECCV_Pose2Mesh,Kocabas_PARE_2021,Kocabas_SPEC_2021,song2020lgd,Bogo:ECCV:2016,kolotouros2021prohmr}, monocular videos~\cite{kanazawa2019learning,kocabas2020vibe,choi2021beyond,sun2019human,luo20203d,zanfir2020weakly,yuan2021simpoe,Zhang:ICCV:2021} or multi-view camera sequences~\cite{huang2017towards,gall2010optimization,joo2018total,saini2019markerless,wang2017outdoor,dong2020motion}.
SPIN~\cite{kolotouros2019learning} estimates SMPL~\cite{loper2015smpl} parameters from single RGB images by combining deep learning with optimization frameworks.
METRO~\cite{lin2021end-to-end} reconstructs human meshes without relying on parametric body models.
Most methods require ``full-body'' images and therefore lack robustness when parts of the body are occluded or truncated, as it is the case with the \interactee{} in egocentric videos. EFT~\cite{joo2020eft} injects crop augmentations at training time to better reconstruct highly truncated people. PARE~\cite{Kocabas_PARE_2021} explicitly learns to predict body-part-guided attention masks. However, these methods exhibit a significant performance drop when applied to egocentric 
data. Our dataset helps fill this performance gap, as we show in Sec.~\ref{sec:experiment}.

The problem of \emph{egocentric} pose estimation is receiving growing attention. 
Most methods estimate the \textit{camera wearer}'s 3D skeleton, based on images, IMU data, scene cues or body-object interactions~\cite{luo2020kinematics,jiang2017seeing,yuan2019ego,shiratori2011motion,tome2020selfpose,tome2019xr,guzov2021human}. 
You2Me~\cite{ng2020you2me} estimates the camera wearer's pose given the \interactee{}'s pose as an additional cue. 
Liu et al.~\cite{liu20204d} estimate 3D human pose and shape of the \interactee{} given egocentric videos in outdoor scenes, with limited interaction diversity.

\myparagraph{Egocentric social interaction learning.}
Egocentric videos provide a unique way to study social interactions.
Most methods focus on social interaction recognition~\cite{fathi2012social,ryoo2013first,narayan2014action,yonetani2016recognizing,li2019deep,dhand2016accuracy,aghaei2016whom,yang2016wearable, aghaei2018towards}.
Lee et al.~\cite{lee2012discovering} produce a storyboard summary of the camera wearer's day given egocentric videos.
Northcutt et al.~\cite{northcutt2020egocom} collect an egocentric communication dataset focusing on conversations.
Recently Ego4D~\cite{ego4d} dataset collects massive egocentric videos for various tasks including hand-object and social interaction understanding, 
making significant advances in stimulating future research in the egocentric domain. \name{} is unique in that we are the first egocentric dataset that provides rich 3D annotations including accurate 3D human pose and shape for all interacting subjects.

\section{Building the \name{} Dataset}
\label{sec:dataset}

\name{} collects sequences capturing subjects performing diverse social interactions in various indoor scenes.
For each sequence, two subjects are involved in one or more interaction scenarios (Sec.~\ref{sec:scenarios}). Their performance is captured from both egocentric- and third-person-views. One subject (the camera wearer) wears a HoloLens2 headset~\cite{hololens2}, capturing multi-modal egocentric data (RGB, depth, head, hand and eye gaze tracking streams). Their interaction partner, \ie \interactee{}, does not wear any device.
The camera wearer's HoloLens2 is calibrated and synchronized with three to five Azure Kinect cameras~\cite{azurekinect} which capture the interaction from different viewpoints (Sec.~\ref{sec:setup}).
Based on this multi-view data, we acquire rich ground-truth annotations for all frames, including 3D full-body pose and shape for both interacting subjects and the reconstructed 3D scene (Sec.~\ref{sec:motion-reconstruct}).
Statistics for \name{} are reported in Sec.~\ref{sec:data_details}.

\subsection{Interaction Scenarios}
\label{sec:scenarios}

To guide the subjects and obtain rich, diverse body motions, we define multiple interaction scenarios within five major interaction categories in sociology studies~\cite{nisbet1970social}: \textit{cooperation}, \textit{social exchange}, \textit{conflict}, \textit{conformity} and \textit{others}, spanning diverse action types (Tab.~\ref{tab:scenarios}) and body poses (Fig.~\ref{fig:qualitative-example}).
For each sequence, we pre-define one or more interaction scenarios and ask the two participants to interact accordingly. 
We allow the subjects to improvise within each interaction scenario to ensure intra-class variation.
The motion diversity is further increased with various human-scene interactions by capturing in 3D scenes.

\begin{table*}[t!]
\centering
\caption{\name{} interaction scenarios.}
\begin{tabular}{ll}
  \toprule[1pt]
  
  \textbf{Category} & \textbf{Interaction Scenarios} \\
  \midrule
  Cooperation  &  Guess by Action game, catching and tossing, searching for items, etc. \\
  Social exchange  &  Teaching to dance/workout, giving a presentation, etc. \\
  Conflict  &  Arguing about a specific topic \\
  Conformity  &  One subject instructs the other to perform a task \\
  Others  &  Haggling, negotiation, promotion, self-introduction, casual chat, etc. \\
  \midrule[1pt]
  
  \multirow{2}{*}{\textbf{Action Types}} 
& Sitting, standing, walking, dancing, exercising, bending, lying, \\
  & grasping, squatting, drinking, passing objects, catching, throwing, etc. \\

  \bottomrule[1pt]
 \end{tabular}
\label{tab:scenarios}
\end{table*}

\subsection{Data Acquisition Setup}
\label{sec:setup}

\begin{figure}[t]
\centering
\includegraphics[width=0.8\linewidth]{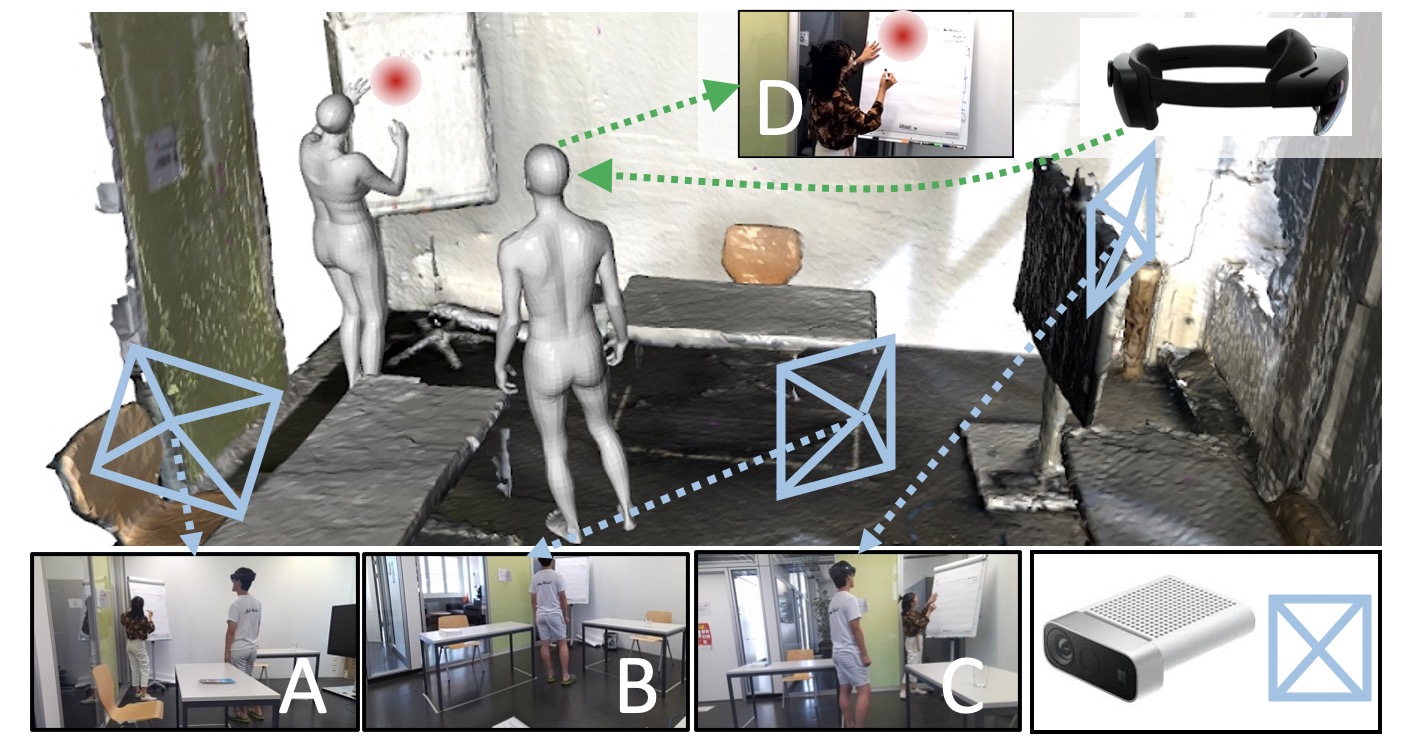}
\caption{Capture setup. Multiple Azure Kinects capture the interactions from different views (A, B, C), and a synchronized HoloLens2 worn by one subject captures the egocentric view image (D), as well as the eye gaze (\textcolor{red}{red} circle) of the camera wearer.}
\label{fig:setup}
\end{figure}

As mentioned above, \name{} collects egocentric- and third-person-view multi-modal data, plus 3D scene reconstructions.
Fig.~\ref{fig:setup} illustrates our system setup.

\myparagraph{Egocentric-view capture.}
We use a Microsoft Hololens2~\cite{hololens2} headset to record egocentric data.
Using the Research Mode API~\cite{hl2_rm}, we capture RGB videos (1920$\times$1080) at 30 FPS, long-throw depth frames (512$\times$512) at 1-5 FPS, 
as well as eye gaze, hand and head tracking at 60 FPS. Note that we do not record depth at a higher framerate (AHAT) due to the ``depth aliasing'' described in~\cite{hl2_rm}. We observe that captures exhibit typical challenges for limited power-devices, like frame drops and blurry images.

\myparagraph{Third-person multi-view capture.}
We use three to five Azure Kinect cameras~\cite{azurekinect} (denoted by \textit{Cam1}$\sim$\textit{Cam5}) to capture multi-view, synchronized RGB-D videos of interacting subjects. Having multi-view data helps our motion reconstruction pipeline for ground-truth acquisition (Sec.~\ref{sec:motion-reconstruct}).
The cameras are fixed during recording. They capture synchronized RGB frames (1920$\times$1080) and depth frames (640$\times$576) at 30 FPS.

\myparagraph{3D scene representation.}
We pre-scan the environment using an iPhone12 Pro Max running the 3D Scanner app~\cite{3dscanner}.
Scene reconstructions are stored as 3D triangulated meshes, each with $10^5\sim10^6$ vertices. 
We choose this procedure for its efficiency and reconstruction quality.

\myparagraph{Calibration and Synchronization.}
For each Kinect, we extract its camera parameters 
via the Azure Kinect DK~\cite{azurekinect}.
For the HoloLens2, we get its camera parameters as exposed by Research Mode~\cite{hl2_rm}.
We synchronize the Kinects via hardware, using audio cables. Since it is not possible to synchronize HoloLens2 and Kinect in a similar way, we use a flashlight visible to all devices as signal for the first frame. 
Kinect-Kinect and Kinect-HoloLens2 cameras are spatially calibrated using a checkerboard and refined by rigid alignment steps (ICP~\cite{icp}). 

The Kinect-HoloLens2 calibration is further optimized based on body keypoints (Sec.~\ref{sec:motion-reconstruct}).
We use \textit{Cam1} to define our world coordinate frame origin. Once we calibrate the HoloLens2 coordinate frame with \textit{Cam1}'s world origin, we can track the headset position, and therefore its cameras, by relying on its built-in head tracker~\cite{hl2_rm}.
We also register the 3D scene into the coordinate frame of \textit{Cam1}, and reconstruct the human body in this space (see details in Supp.~Mat.).

\begin{figure}[t]
\begin{minipage}[b]{0.48\linewidth}
\centering
\includegraphics[width=0.9\linewidth]{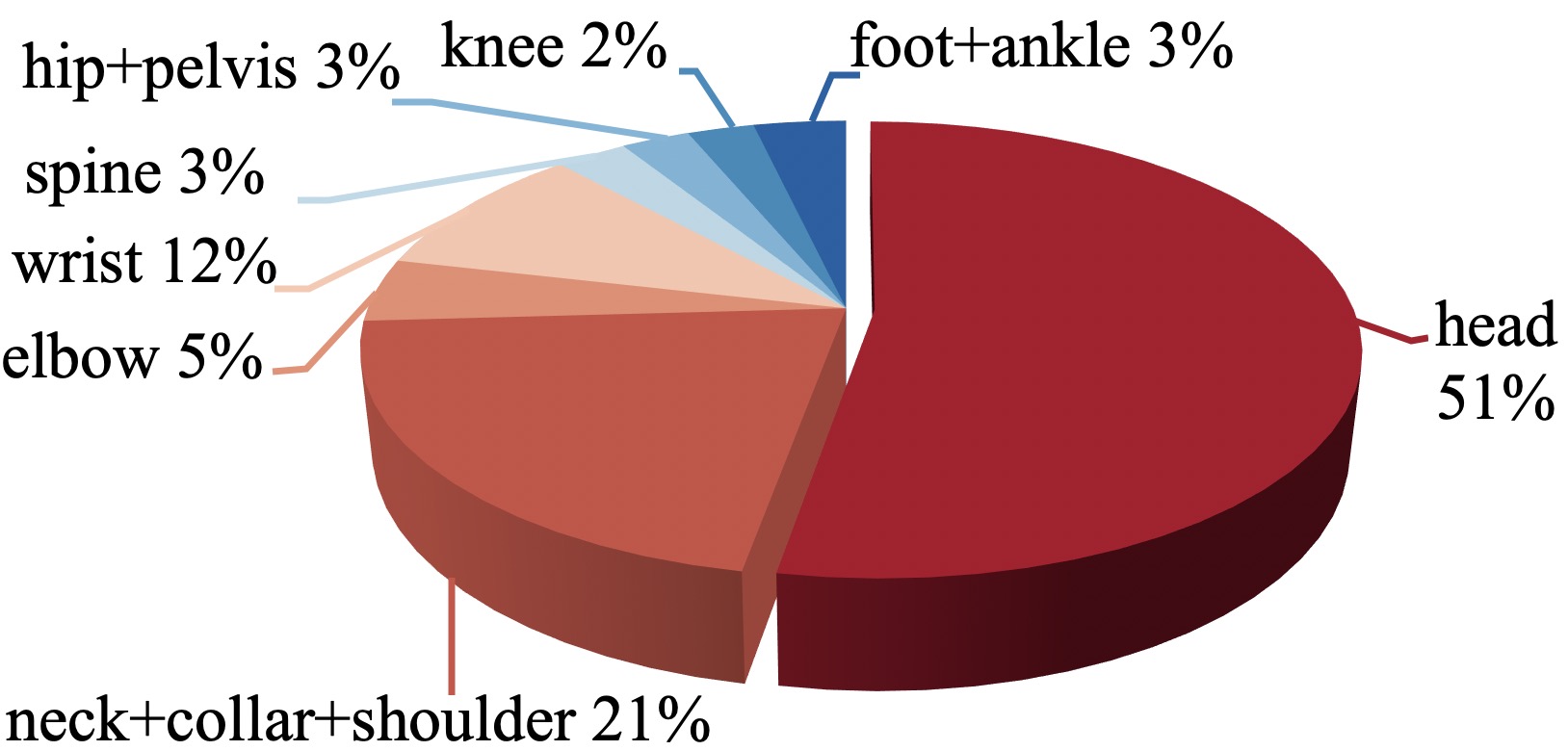}
\caption{\textbf{Which body part attracts more attention?}
  For each joint group, \% of the occurrences it is the closest to the 2D gaze point in the image.}
\label{fig:gaze_pie_chart}
\end{minipage}
\begin{minipage}[b]{0.5\linewidth}
\centering
\includegraphics[width=\linewidth]{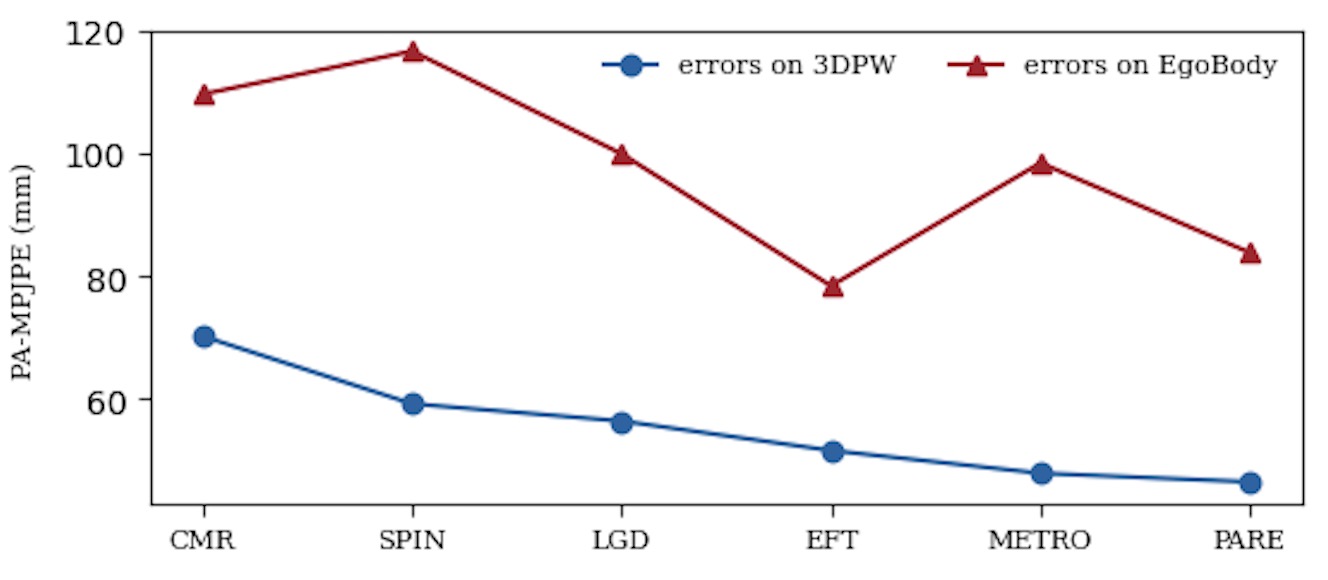}
\caption{Accuracy of SoTAs on 3DPW and \name{} with the advance of the 3DHPS field.}
\label{fig:error_curve_3dpw_vs_ours}
\end{minipage}
\end{figure}

\subsection{Ground-truth Acquisition}
\label{sec:motion-reconstruct}

Given the RGB-D frames captured with the multi-Kinect rig and the egocentric frames, our motion reconstruction pipeline estimates, for each frame and each subject, the corresponding SMPL-X body parameters~\cite{pavlakos2019expressive}, including the global translation $\boldsymbol{\gamma} \in \mathbb{R}^3$, body shape $\boldsymbol{\beta} \in \mathbb{R}^{10}$, pose $\boldsymbol{\theta} \in \mathbb{R}^{96}$ (body and hand) and facial expression $\boldsymbol{\phi} \in \mathbb{R}^{10}$. 
To address the challenges posed by not perfectly accurate factory calibration of HoloLens2 depth, which further leads to inaccurate Kinect-HoloLens2 calibration, we propose a keypoint-based refinement scheme to better leverage observations from the HoloLens2.
We introduce the first solution to reconstruct accurate 3D human pose, shape and motions with multi-view Kinect cameras and an embodied HMD. 
Thanks to the refined Kinect-HoloLens2 calibration, this provides accurate per-frame pose, natural human motion dynamics and realistic human-scene interactions for both egocentric- and third-person-view frames.
Note that we estimate the body in the coordinate frame of \textit{Cam1}.

\myparagraph{Data preprocessing.}
We use OpenPose~\cite{openpose} to detect 2D body joints in all (Kinect and HoloLens2) RGB frames. OpenPose identifies people in the same image by assigning a body index to each detected person. In general, this works well, but gives false positives which we process afterwards.
To extract human body point clouds from Kinect depth frames, we use Mask-RCNN~\cite{he2017mask} and DeepLabv3~\cite{he2017mask}. 
We manually inspect the data to remove spurious detections (\eg irrelevant people in the background, and scene objects misdetected as people). We also ensure consistent subject identification across frames and views, and manually fix inaccurate 2D joint detections, mostly due to body-body and body-scene occlusions.
See Supp.~Mat. for more details.

\myparagraph{Per-frame fitting.}
As in~\cite{Zhang:ICCV:2021}, 
given Kinect depth and 2D joints, we first optimize the SMPL-X parameters for each subject/frame separately, minimizing an objective function similar to that defined in~\cite{hassan2019resolving}:
\begin{equation}
\begin{split}
    E(\boldsymbol{\beta}, \boldsymbol{\gamma}, \boldsymbol{\theta}, \boldsymbol{\phi}) &= 
    E_J + \lambda_D E_D + E_{prior} + \lambda_{contact} E_{contact} + \lambda_{coll} E_{coll},
\end{split}
\end{equation}
where $\boldsymbol{\beta}$, $\boldsymbol{\gamma}, \boldsymbol{\theta}, \boldsymbol{\phi}$ are optimized SMPL-X parameters.

Given the preprocessed OpenPose 2D joints 
$J_{OP}^v$ from $n$ views ($v \in \{1,...,n\}$),
the multi-view joint error term $E_J$ minimizes the sum of 2D distances between $J_{OP}^v$ and the 2D projection of SMPL-X joints onto camera view $v$ for all views:
\begin{equation}
    E_J(\boldsymbol{\beta}, \boldsymbol{\gamma}, \boldsymbol{\theta}, \boldsymbol{\phi}) = \sum_{view\ v}  E_{J_v}(\boldsymbol{\beta}, \boldsymbol{\gamma}, \boldsymbol{\theta}, \boldsymbol{\phi}, J_{OP}^v, K_v, T_v),
\end{equation}
where $K_v$ denotes the intrinsics parameters of camera $v$, and $T_v$ denotes the extrinsics between \textit{Cam} $v$ and \textit{Cam 1}. 
The depth term $E_D$ penalizes discrepancies between the estimated body surface and body depth point clouds for all views; 
$E_{prior}$ represents body pose, shape and expression priors;
$E_{contact}$ encourages scene-body contacts; and $E_{coll}$ penalizes scene-body collisions. 
The $\lambda_i$s weight the contribution of each term. 
We refer the reader to~\cite{hassan2019resolving,Zhang:ICCV:2021} for more details.

\myparagraph{Kinect-HoloLens2 calibration refinement.}
The Kinect-Hololens2 calibration is represented by the extrinsics $T$ between Kinect \textit{Cam1}, and the HoloLens2 coordinate system's origin.
For each capture session, this origin is fixed in the world~\cite{hl2_rm}; as the HMD moves, its head tracker provides the transformation between this origin and each egocentric frame $t$, denoted by $T^{ego}_t$.
To address the inaccurate initial Kinect-HoloLens2 calibration $T_{init}$ caused by imperfect HoloLens2 depth factory calibration,
we propose a keypoint-based scheme to refine it. 
For each frame $t$, we project the 3D SMPL-X joints $J_{3D, t}$ (obtained from per-frame fitting, in \textit{Cam1}'s coordinate) onto the egocentric image. We minimize the 2D error between the projected 2D joints and the OpenPose joint detections $J_{OP, t}^{ego}$ of the egocentric frame $t$, and optimize the transformation $T$:

\begin{equation}
    E_T (T) = \sum_{t} ||K^{ego} T_t^{ego} T J_{3D, t} - J_{OP, t}^{ego}||_2^2 + \lambda ||T - T_{init}||_2^2,
\end{equation}
where $K^{ego}$ denotes the HoloLens2 RGB camera intrinsic parameters, and $\lambda$ weights the regularizer. 

\myparagraph{Temporally consistent fitting.}
Per-frame fitting gives us a set of reasonable, initial pose estimates, which however are jittery and inconsistent over time.
We therefore run a second optimization stage based on LEMO priors~\cite{Zhang:ICCV:2021} to obtain smooth, realistic human motions.
Furthermore, to improve consistency between egocentric- and third-person-view estimates, we consider also egocentric data given the refined Kinect-HoloLens2 calibration. We take OpenPose 2D joint estimations from HoloLens2 RGB frames and use them as further constraints. 
Still, we optimize for each subject separately.
The resulting objective function minimized in the temporal fitting stage is:
\begin{equation}
\begin{split}
    E(\boldsymbol{\gamma}, \boldsymbol{\theta}, \boldsymbol{\phi}) &= 
    E_J + E_{J_{ego}} + E_{prior} + \lambda_{smooth} E_{smooth} + \lambda_{fric} E_{fric} ,
\end{split}
\end{equation}
where $E_{fric}$ is the contact friction term defined in~\cite{Zhang:ICCV:2021} to prevent body sliding, $E_{smooth}$ and $E_{prior}$ denote temporal and static priors as in~\cite{Zhang:ICCV:2021}. $E_{J_{ego}}$ is the 2D projection term which minimizes the error between OpenPose detections on egocentric view frames and the 2D projections of SMPL-X joints onto the egocentric view; $E_{J_{ego}}$ is only enabled for the \interactee{} when they are visible in the egocentric frames. The $\lambda_i$s weight balance the contribution of each term.

\section{\name{} Dataset}
\label{sec:data_details}

\begin{figure*}[t]
    \centering
    \includegraphics[width=0.85\linewidth]{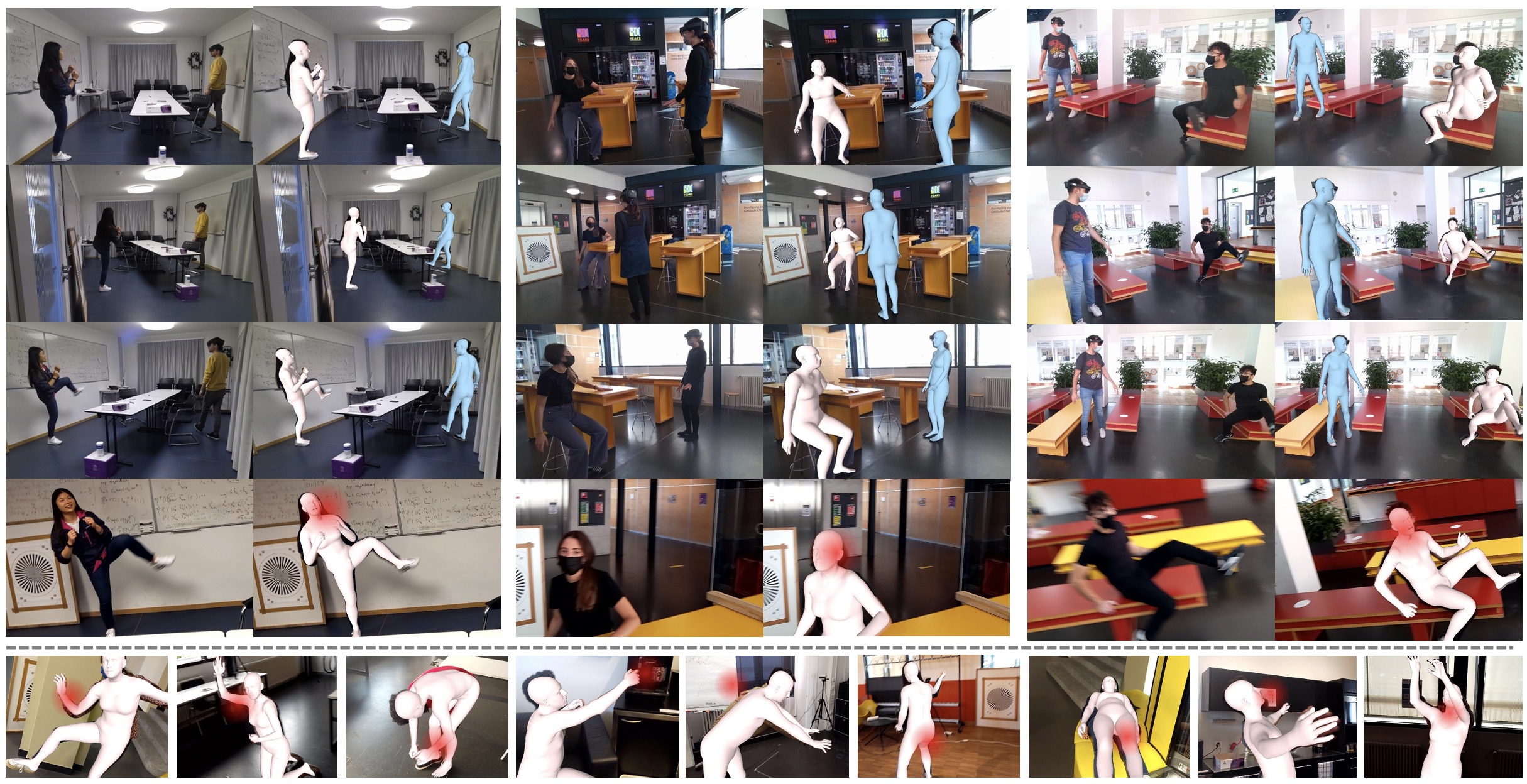}
    \caption{Reconstructed ground-truth bodies overlaid on third-person-view images from 3 Kinects (row 1-3), and the corresponding egocentric view image (row 4). Left/middle/right shows three different frames. Row 5 shows more examples from the egocentric view. \textcolor{blue}{Blue} denotes the camera wearer, and \textcolor{pink}{pink} denotes the \interactee{}. Eye gaze of the camera wearer are in \textcolor{red}{red} circles.}
    \label{fig:qualitative-example}
\end{figure*}

\name{} collects 125 sequences from 36 subjects (18 male and 18 female) performing diverse social interactions in 15 indoor scenes.
In total, there are 219,731 synchronized frames captured from Azure Kinects, from multiple third-person-views. We refer to this as the ``Multi-view (MV)Set''. For each MV frame, we provide 3D human full-body pose and shape annotations (as SMPL-X parameters) for both interacting subjects together with the 3D scene mesh.
Furthermore, we have 199,111 egocentric RGB frames (the ``EgoSet''), captured from HoloLens2, calibrated and synchronized with Kinect frames. Given the camera wearer's head motion, the \interactee{} is not visible in every egocentric frame; in total, we have 175,611 frames with the \interactee{} visible in the egocentric view (``EgoSet-\interactee{}''). 
Fig.~\ref{fig:qualitative-example} shows example images. 
For EgoSet, we also collect the head, hand and eye tracking data, plus the depth frames from the HoloLens2. 
We also provide SMPL~\cite{loper2015smpl} body annotations via the official transfer tool~\cite{smplx2smpl}. Below we provide dataset statistics; for more detailed analysis and ground truth annotation quality please refer to the Supp.~Mat.

\myparagraph{Training/validation/test splits.} We split data into training, validation and test sets such that they have no overlapping subjects. 
The \name{} training set contains 116,630 MVSet frames, 105,388 EgoSet frames and 90,124 EgoSet-\interactee{} frames. 
The \name{} validation set contains 29,140 MVSet frames, 25,416 EgoSet frames and 23,332 EgoSet-\interactee{} frames. 
The test set contains 73,961 MV frames, 68,307 EgoSet frames and 62,155 Ego-\interactee{} frames.

\myparagraph{Joint visibility.}
The camera wearer's motion, the headset's field of view and the close distance between the interacting subjects cause the \interactee{} to be often truncated in the egocentric view. 
To quantify the occurrences of truncations, we project the fitted 3D body joints onto the HoloLens2 images, and deem a projected 2D joint as ``visible'' if it lies inside the image. 
As shown in Fig.~\ref{fig:experiment_analysis} (2nd row, right), the lower body parts are more frequently truncated in the images.
Please refer to Sec.~\ref{sec:experiment} for the impact of joint visibility on 3DHPS estimation performance.

\myparagraph{Eye gaze and attention.}
We can combine the HoloLens2 eye gaze tracking with our 3D reconstruction of the scene/people to estimate the 3D location the user looks at, and project it on the egocentric images (interpreted as where the user's ``attention'' is focused),
thereby obtaining valuable data to understand interactions. 
We observe that the camera wearer's attention is highly focused on the \interactee{} during interactions. and tends to be closer to the upper body joints (Fig.~\ref{fig:gaze_pie_chart}), which in turn results in lower visibility for the lower body parts.

\section{Experiments}
\label{sec:experiment}
We leverage \name{} to introduce the first benchmark for 3D human pose and shape (3DHPS) estimation from egocentric images.
Given a single RGB image of a target subject, the goal of a 3DHPS method is to estimate a human body mesh and a set of camera parameters, which best explain the image data. 
State-of-the-art (SoTA) 3DHPS methods are mostly trained and evaluated on third-person-view data, and their performance is starting to saturate on common third-person-view datasets~\cite{von2018recovering,ionescu2013human3} (see Fig.~\ref{fig:error_curve_3dpw_vs_ours}); yet, their capabilities to generalize to real-world scenarios (\eg cropped or blurry images) are still limited~\cite{Patel:CVPR:2021}.
With \name{}, we can test their capabilities on egocentric images.

We define a benchmark for 3DHPS methods on our EgoSet-\interactee{} test set. 
Within the social interaction scenarios, the input will be an egocentric view image of the \interactee{}. 
We evaluate SoTA methods and show that their performance significantly drops on our data.
We expose limitations of existing methods by in-depth analysis (Sec.~\ref{sec:baseline_eval}), given that the egocentric view brings considerable challenges 
that are rarely present in existing third-person-view datasets.

We also provide valuable insights to boost their performance for egocentric scenarios.
In particular, we show that our EgoSet-\interactee{} training set can help address the challenges brought by egocentric view data: using it, we fine-tune three recent methods, SPIN~\cite{kolotouros2019learning}, METRO~\cite{lin2021end-to-end} and EFT~\cite{joo2020eft}, achieving significantly improved accuracy and robustness on both our test set (Sec.~\ref{sec:baseline_impr}) and over a cross-dataset evaluation on the You2Me~\cite{ng2020you2me} dataset (Sec.~\ref{sec:baseline_impr_you2me}).

\subsection{Benchmark Evaluation Metrics}\label{sec:metrics}
We employ two common metrics: \textbf{Mean Per-Joint Position Error (MPJPE)} and \textbf{Vertex-to-Vertex (V2V)} errors. 
We use two types of alignments before computing the accuracy for each metric: (1)~translation-only alignment (aligns the bodies at the pelvis joint~\cite{Patel:CVPR:2021}) and (2)~Procrustes Alignment~\cite{gower1975generalized} (``PA'', solves for scale, translation and rotation). 
Results are by default reported with translation-only alignment unless specified with the ``PA-'' prefix. 
\textbf{MPJPE} is the mean Euclidean distance between predicted and ground-truth 3D joints, evaluated on 24 SMPL body joints. 
\textbf{V2V} error is the mean Euclidean distance over all body vertices, computed between two meshes. 

\begin{table}[tb]
\centering
\caption{Evaluation of SoTA 3DHPS estimation methods on our test set. All metrics are in {\em mm}. ``PA-'' stands for Procrustes alignment. ``SPIN-ft'', ``METRO-ft'' and ``EFT-ft'' denote results of fine-tuning SPIN, METRO and EFT on our training set.}
\label{tab:baseline_eval}
\begin{tabular}{lcccccc}
\toprule[1pt]
Method                              & MPJPE~$\downarrow$ & PA-MPJPE~$\downarrow$& V2V~$\downarrow$& PA-V2V~$\downarrow$ \\
\midrule
CMR~\cite{kolotouros2019cmr}         & 200.7   & 109.6   & 218.7 & 136.8 \\
SPIN~\cite{kolotouros2019learning}   & 182.8   & 116.6   & 187.3 & 123.7 \\
LGD~\cite{song2020lgd}               & 158.0   & 99.9    & 168.3 & 106.0 \\
METRO~\cite{lin2021end-to-end}       & 153.1   & 98.4    & 164.6 & 106.4 \\
PARE~\cite{Kocabas_PARE_2021}        & 123.0   & 83.8    & 131.4 & 89.7  \\
EFT~\cite{joo2020eft}                & 123.9   & 78.4    & 135.0 & 86.0  \\
\hline
SPIN-ft (Ours)                       &  106.5  & 67.1    & 120.9 &  78.3     \\
METRO-ft (Ours)                      &  \textbf{98.5}   &  66.9   & \textbf{110.5} & 76.8 \\
EFT-ft (Ours)                        &  102.1  & \textbf{64.8}   & 116.1 & \textbf{74.8} \\
\bottomrule[1pt] 
\end{tabular}
\end{table}

\subsection{Baseline Evaluation}\label{sec:baseline_eval}
Tab.~\ref{tab:baseline_eval} summarizes the evaluation of SoTA 3DHPS methods from different categories: (1)~fitting-based method~\cite{song2020lgd}; and regression-based methods that (2)~predict parameters of a parametric body model~\cite{kolotouros2019cmr,kolotouros2019learning,Kocabas_PARE_2021,joo2020eft} or (3) predict non-parametric body meshes~\cite{lin2021end-to-end}.
For each baseline method, we use the best performing model provided by the authors (trained with the optimal training data).

In Fig.~\ref{fig:error_curve_3dpw_vs_ours} we plot the PA-MPJPE error of these methods on our dataset and on an existing major third-person-view benchmark\footnote{The results on 3DPW are taken from the respective original papers.}, 
On average, the methods 
yield a 77\% higher 3D joint error on \name{} than on 3DPW. More importantly, while the accuracy curve drives towards saturation on 3DPW, different SoTA methods still show largely varying performance on our dataset. 
This suggests that current datasets are not sufficient to train models that can handle egocentric view images well. 
Below we discuss two key challenging factors that impact performance.

\begin{figure}[tb]
\centering
    \begin{minipage}[b]{\linewidth}
    \centering  
    {
    \includegraphics[width=0.98\linewidth]{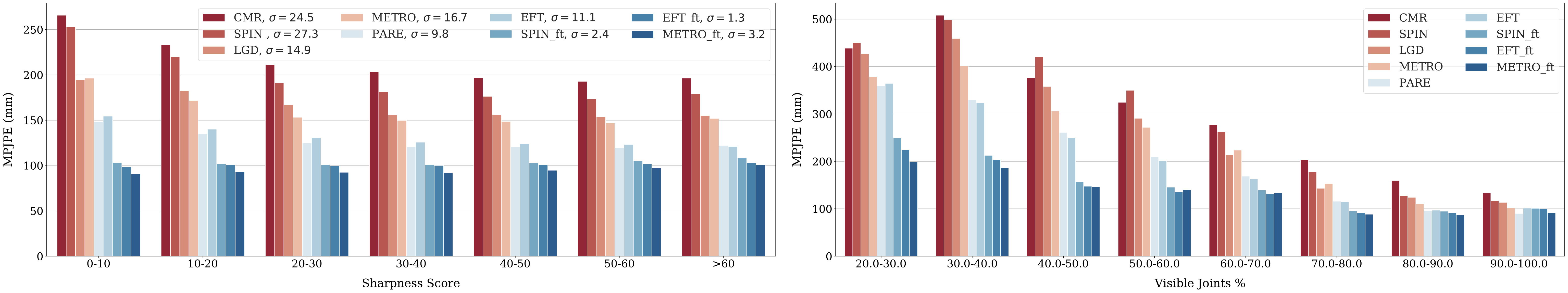}
    }
    \end{minipage}
    
    \begin{minipage}[b]{0.49\linewidth}
    \centering  
    {
    \includegraphics[width=\linewidth]{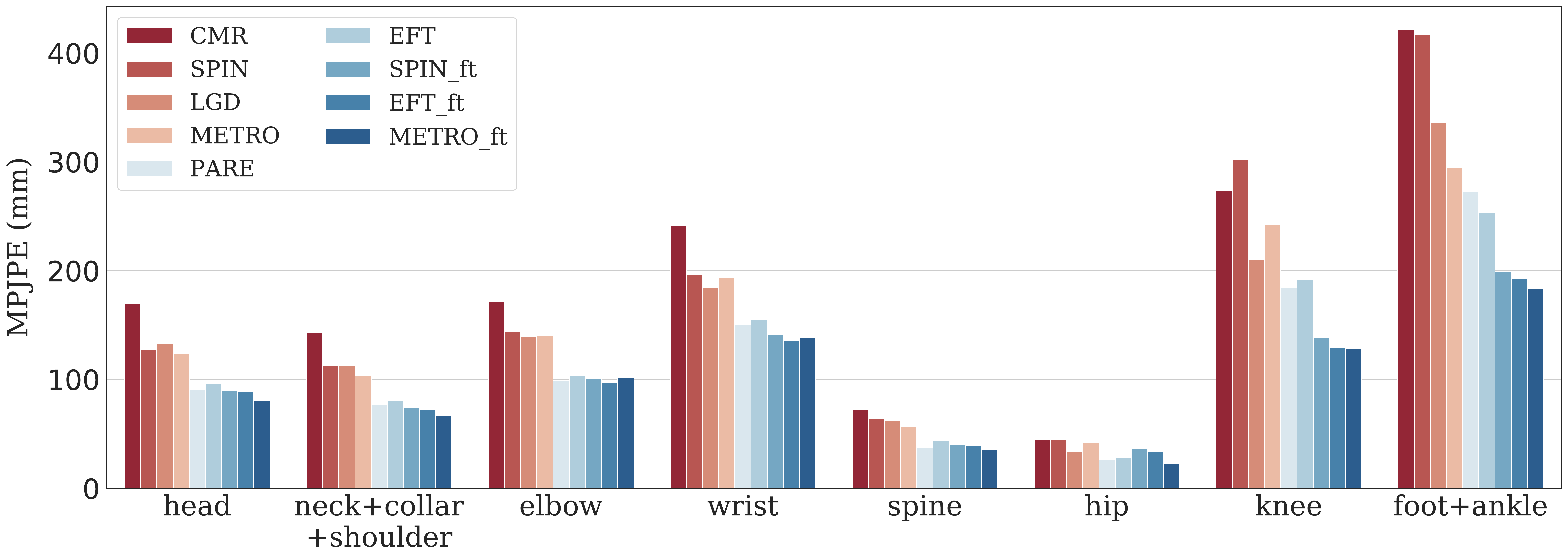}
    }
    \end{minipage}
    \begin{minipage}[b]{0.49\linewidth}
    \centering  
    {
    \includegraphics[width=\linewidth]{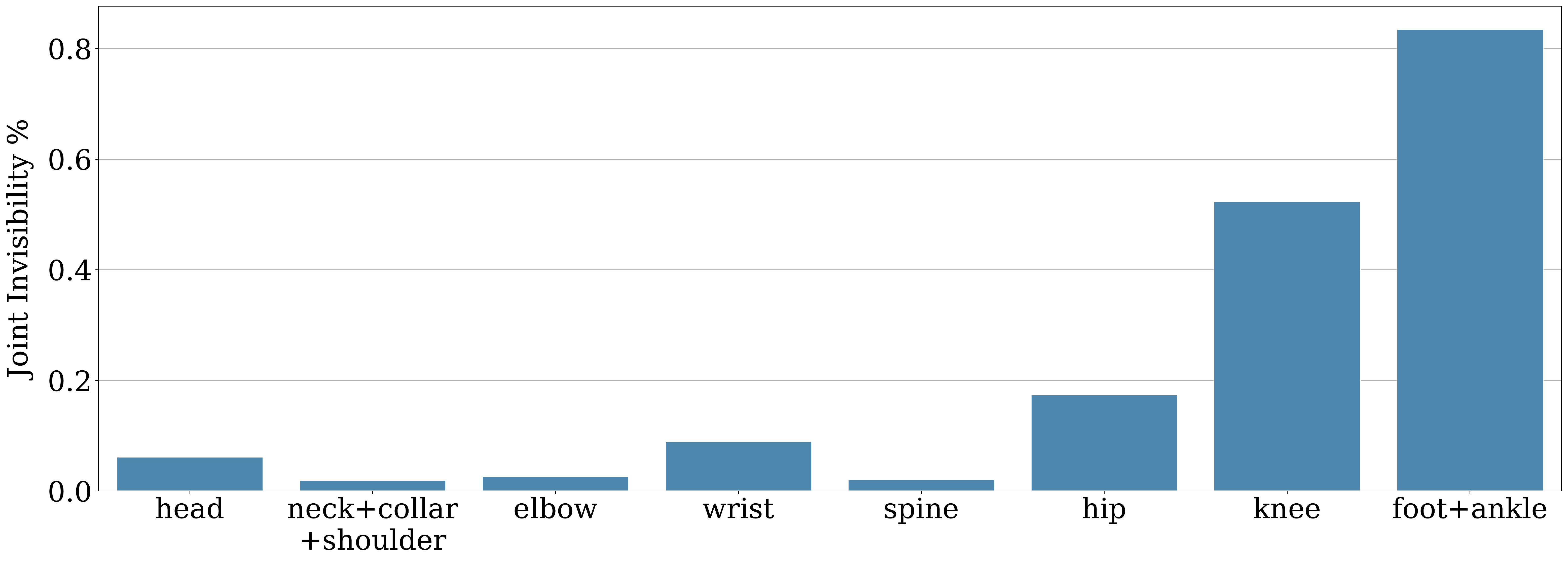}
    }
    \end{minipage}
\caption{First Row: Impact of motion blur (left) / joint visibility (right) on SoTA method accuracies.
Second Row: 3D joint error analysis by body parts (left) / ratio of each joint group being \textit{in}visible (truncated) from the images in our test set (right).
}
\label{fig:experiment_analysis}
\end{figure}

\myparagraph{Motion blur.}
Motion blur is common in the egocentric view images due to the motion of the camera wearer. 
To study how motion blur influences 3DHPS estimation accuracy, we plot in Fig.~\ref{fig:experiment_analysis} (1st row, left) the MPJPE of all methods vs. the image sharpness score.
The sharpness score is defined as the variance of the Laplacian of an image~\cite{pech2000diatom}, upper-thresholded at 60; higher scores mean sharper images. 
We observe that, surprisingly, most methods are insensitive to blurriness, except for heavily blurred cases (score \textless 10). 
However, our fine-tuned models (SPIN-ft / METRO-ft / EFT-ft ) are more robust against motion blur: among all methods, they achieve the lowest standard deviation over the seven image sharpness levels; see the number next to each method in the legend of Fig.~\ref{fig:experiment_analysis} (1st row, left).

\myparagraph{Joint visibility.}
While most 3DHPS methods assume that the target body is (almost) fully visible in the image as in existing third-person-view datasets such as 3DPW~\cite{von2018recovering}, this is seldom the case in egocentric view images.
To assess the importance of this issue, we analyze the performance of each baseline with respect to the portion of visible body joints (``visibility'', see Sec.~\ref{sec:data_details}) in the images from our test set. The result is summarized in Fig.~\ref{fig:experiment_analysis} (row 1, right).
Note that our definition of a joint's visibility is related to, but differs from, the concept of \textit{occlusion}: both measure how much pixel information is missing for a body part, but visibility focuses on how much of the body is \textit{truncated} from the image. A joint that is occluded by an object can still be considered visible by our definition.

Overall, all methods yield a lower error when there is less body truncation. 
Two recent methods, PARE~\cite{Kocabas_PARE_2021} and EFT~\cite{joo2020eft}, achieve the best results. 
PARE is designed to be robust against occlusions by explicitly employing a body part attention mechanism, whereas EFT handles body truncation ``implicitly'' by aggressively cropping images as training data augmentation.

We further plot the MPJPE and the \textit{in}visibility ratio of each joint group in Fig.~\ref{fig:experiment_analysis} (2nd row). Overall the two are in accordance: the lesser a joint is visible, the higher the error it exhibits. 
An exception is on the wrist joints: despite good visibility, their error remains relatively high. 
As observed also in \cite{Kocabas_PARE_2021}, high errors on the extremities are a common problem with existing 3DHPS models, possibly because most current models only use a single, global feature from the input image for regression.
This points to potential future work that deploys local image features, which has been shown effective in recent 3DHPS models~\cite{Kocabas_PARE_2021,guler2019holopose}.
%

\subsection{Baseline Improvement}\label{sec:baseline_impr}
To evaluate the effectiveness of the \name{} training set, we use it to fine-tune three of the baseline methods: two model-based methods, SPIN~\cite{kolotouros2019learning} and EFT~\cite{joo2020eft}, as they both use the same architecture (HMR~\cite{kanazawa2018end} network) that is the backbone for many other recent models~\cite{kocabas2020vibe,kanazawa2018end,rong2021frankmocap}; a model-free method METRO~\cite{lin2021end-to-end} which directly predicts the body mesh.
The pre-trained EFT differs from SPIN majorly in that it is trained with extended 3D pseudo ground-truth data (from the EFT-dataset) and uses aggressive image cropping as data augmentation.
We use the same hyperparameters provided by the authors and select the fine-tuned model with the best validation score.

As shown in Tab.~\ref{tab:baseline_eval}, after fine-tuning, the error is largely reduced for all three methods on all metrics: 
SPIN-ft/METRO-ft/EFT-ft has 42\%/36\%/18\% lower MPJPE, and 35\%/33\%/14\% V2V than their corresponding original models.

The improvement can also be seen for all blurriness/visibility categories in Figs.~\ref{fig:experiment_analysis}.
For the motion blur specifically, the fine-tuned models not only achieve a lower error at every image sharpness level, but also show increased robustness. 
This is shown by the standard deviations of each method across the sharpness levels, dropping from 27.3 to 2.4 for SPIN, from 16.7 to 3.2 for METRO, and from 11.1 to 1.3 for EFT, respectively, after fine-tuning.  
The results show that our training set can serve as an effective source to adapt existing 3DHPS methods to the egocentric setting.

\subsection{Cross-dataset Evaluation on You2Me}\label{sec:baseline_impr_you2me}
Is the effect of our training set only specific to our capture scenario, or does it generalize to other egocentric pose estimation datasets?
To verify this, 
we evaluate SPIN, EFT and METRO against their fine-tuned counterparts on the You2Me~\cite{ng2020you2me} dataset. Here we report the PA-MPJPE for pose errors (in \textit{mm}): SPIN (152.8) vs.~SPIN-ft (87.9); EFT (95.8) vs.~EFT-ft (85.6), and METRO (117.7) vs.~METRO-ft (88.2). Again, fine-tuning on our training set improves all models' performance; see Supp.~Mat. for more details.
These results suggest that our data empowers existing models with the ability to address challenges faced in the generic egocentric view setup. 

\section{Conclusion}
\label{sec:conclusion}

We presented \name{}, a dataset capturing human pose, shape and motions of interacting people in diverse environments. \name{} collects multi-modal egocentric- and third-person-view data, accompanied by ground-truth 3D human pose and shape for all interacting subjects.
With this dataset, we introduced a benchmark on egocentric-view 3D human body pose and shape (3DHPS) estimation, systematically evaluated and analyzed limitations of state-of-the-art methods on the egocentric setting, and demonstrated a significant, generalizable performance gain in them with the help of our annotations. 
This paper has shown \name{}'s unique value for the 3DHPS estimation task, and we see its great potential in moving the fields towards a better understanding of egocentric human motions, behaviors, and social interactions.
In the future, adding more participants and even richer data modalities (\eg audio recordings and motion annotations by natural language descriptions) could further enrich the dataset. 

{
\myparagraph{Acknowledgements.} 
This work was supported by the SNF grant 200021 204840 and Microsoft Mixed Reality \& AI Zurich Lab PhD scholarship.
Qianli Ma is partially funded by the Max Planck ETH Center for Learning Systems. 
We sincerely thank Francis Engelmann, Korrawe Karunratanakul, Theodora Kontogianni, Qi Ma, Marko Mihajlovic, Sergey Prokudin, Matias Turkulainen, Rui Wang , Shaofei Wang and Samokhvalov Vyacheslav for helping with the data capture and processing, Xucong Zhang for the discussion of data collection and Jonas Hein for the discussion of the hardware setup.
}

%
%
\bibliographystyle{splncs04}
\bibliography{egbib}

\begin{thebibliography}{100}
\providecommand{\url}[1]{\texttt{#1}}
\providecommand{\urlprefix}{URL }
\providecommand{\doi}[1]{https://doi.org/#1}

\bibitem{azurekinect}
{Azure Kinect}. \url{https://docs.microsoft.com/en-us/azure/kinect-dk/}

\bibitem{3dscanner}
{Laan Labs 3D Scanner app}.
  \url{https://apps.apple.com/us/app/3d-scanner-app/id1419913995}

\bibitem{hololens2}
{Microsoft Hololens2}. \url{https://www.microsoft.com/en-us/hololens}

\bibitem{smplx2smpl}
{SMPL model transfer}.
  \url{https://github.com/vchoutas/smplx/tree/master/transfer_mode}

\bibitem{agarwal2005recovering}
Agarwal, A., Triggs, B.: Recovering 3d human pose from monocular images. IEEE
  transactions on pattern analysis and machine intelligence  \textbf{28}(1),
  44--58 (2005)

\bibitem{aghaei2018towards}
Aghaei, M., Dimiccoli, M., Ferrer, C.C., Radeva, P.: Towards social pattern
  characterization in egocentric photo-streams. Computer Vision and Image
  Understanding  \textbf{171},  104--117 (2018)

\bibitem{aghaei2016whom}
Aghaei, M., Dimiccoli, M., Radeva, P.: With whom do i interact? detecting
  social interactions in egocentric photo-streams. In: 2016 23rd International
  Conference on Pattern Recognition (ICPR). pp. 2959--2964. IEEE (2016)

\bibitem{nisbet1970social}
A.Nisbet, R.: The Social Bond: An Introduction to the Study of Society (1970)

\bibitem{bualan2008naked}
B{\u{a}}lan, A.O., Black, M.J.: The naked truth: Estimating body shape under
  clothing. In: European Conference on Computer Vision. pp. 15--29. Springer
  (2008)

\bibitem{bambach2015lending}
Bambach, S., Lee, S., Crandall, D.J., Yu, C.: Lending a hand: Detecting hands
  and recognizing activities in complex egocentric interactions. In:
  Proceedings of the IEEE international conference on computer vision. pp.
  1949--1957 (2015)

\bibitem{icp}
Besl, P., McKay, N.D.: A method for registration of 3-d shapes. IEEE
  Transactions on Pattern Analysis and Machine Intelligence  \textbf{14}(2),
  239--256 (1992)

\bibitem{Bogo:ECCV:2016}
Bogo, F., Kanazawa, A., Lassner, C., Gehler, P., Romero, J., Black, M.J.: Keep
  it {SMPL}: Automatic estimation of {3D} human pose and shape from a single
  image. In: European Conference on Computer Vision. pp. 561--578 (2016)

\bibitem{openpose}
{Cao}, Z., {Hidalgo Martinez}, G., {Simon}, T., {Wei}, S., {Sheikh}, Y.A.:
  Openpose: Realtime multi-person 2d pose estimation using part affinity
  fields. IEEE Transactions on Pattern Analysis and Machine Intelligence
  (2019)

\bibitem{choi2021beyond}
Choi, H., Moon, G., Chang, J.Y., Lee, K.M.: Beyond static features for
  temporally consistent 3d human pose and shape from a video. In: Proceedings
  of the IEEE/CVF Conference on Computer Vision and Pattern Recognition. pp.
  1964--1973 (2021)

\bibitem{Choi_2020_ECCV_Pose2Mesh}
Choi, H., Moon, G., Lee, K.M.: Pose2mesh: Graph convolutional network for 3d
  human pose and mesh recovery from a 2d human pose. In: European Conference on
  Computer Vision (ECCV) (2020)

\bibitem{damen2018scaling}
Damen, D., Doughty, H., Farinella, G.M., Fidler, S., Furnari, A., Kazakos, E.,
  Moltisanti, D., Munro, J., Perrett, T., Price, W., et~al.: Scaling egocentric
  vision: The epic-kitchens dataset. In: Proceedings of the European Conference
  on Computer Vision (ECCV). pp. 720--736 (2018)

\bibitem{damen2022rescaling}
Damen, D., Doughty, H., Farinella, G.M., Furnari, A., Kazakos, E., Ma, J.,
  Moltisanti, D., Munro, J., Perrett, T., Price, W., et~al.: Rescaling
  egocentric vision: collection, pipeline and challenges for epic-kitchens-100.
  International Journal of Computer Vision  \textbf{130}(1),  33--55 (2022)

\bibitem{dhand2016accuracy}
Dhand, A., Dalton, A.E., Luke, D.A., Gage, B.F., Lee, J.M.: Accuracy of
  wearable cameras to track social interactions in stroke survivors. Journal of
  Stroke and Cerebrovascular Diseases  \textbf{25}(12),  2907--2910 (2016)

\bibitem{dong2020motion}
Dong, J., Shuai, Q., Zhang, Y., Liu, X., Zhou, X., Bao, H.: Motion capture from
  internet videos. In: European Conference on Computer Vision. pp. 210--227.
  Springer (2020)

\bibitem{fang2021reconstructing}
Fang, Q., Shuai, Q., Dong, J., Bao, H., Zhou, X.: Reconstructing 3d human pose
  by watching humans in the mirror. In: Proceedings of the IEEE/CVF Conference
  on Computer Vision and Pattern Recognition. pp. 12814--12823 (2021)

\bibitem{fathi2012social}
Fathi, A., Hodgins, J.K., Rehg, J.M.: Social interactions: A first-person
  perspective. In: 2012 IEEE Conference on Computer Vision and Pattern
  Recognition. pp. 1226--1233. IEEE (2012)

\bibitem{fathi2011understanding}
Fathi, A., Farhadi, A., Rehg, J.M.: Understanding egocentric activities. In:
  2011 international conference on computer vision. pp. 407--414. IEEE (2011)

\bibitem{fieraru2020three}
Fieraru, M., Zanfir, M., Oneata, E., Popa, A.I., Olaru, V., Sminchisescu, C.:
  Three-dimensional reconstruction of human interactions. In: Proceedings of
  the IEEE/CVF Conference on Computer Vision and Pattern Recognition. pp.
  7214--7223 (2020)

\bibitem{gall2010optimization}
Gall, J., Rosenhahn, B., Brox, T., Seidel, H.P.: Optimization and filtering for
  human motion capture. International journal of computer vision
  \textbf{87}(1-2), ~75 (2010)

\bibitem{gower1975generalized}
Gower, J.C.: Generalized procrustes analysis. Psychometrika  \textbf{40}(1),
  33--51 (1975)

\bibitem{grauman2003inferring}
Grauman, K., Shakhnarovich, G., Darrell, T.: Inferring 3d structure with a
  statistical image-based shape model. In: ICCV. vol.~3, p.~641 (2003)

\bibitem{ego4d}
Grauman, K., et~al.: Ego{4D}: {A}round the world in 3000 hours of egocentric
  video. arXiv preprint arXiv:2110.07058  (2021)

\bibitem{guler2019holopose}
Guler, R.A., Kokkinos, I.: Holopose: Holistic 3d human reconstruction
  in-the-wild. In: Proceedings of the IEEE Conference on Computer Vision and
  Pattern Recognition. pp. 10884--10894 (2019)

\bibitem{guzov2021human}
Guzov, V., Mir, A., Sattler, T., Pons-Moll, G.: Human poseitioning system
  (hps): 3d human pose estimation and self-localization in large scenes from
  body-mounted sensors. In: Proceedings of the IEEE/CVF Conference on Computer
  Vision and Pattern Recognition. pp. 4318--4329 (2021)

\bibitem{hassan2019resolving}
Hassan, M., Choutas, V., Tzionas, D., Black, M.J.: Resolving 3d human pose
  ambiguities with 3d scene constraints. In: Proceedings of the IEEE/CVF
  International Conference on Computer Vision. pp. 2282--2292 (2019)

\bibitem{he2017mask}
He, K., Gkioxari, G., Doll{\'a}r, P., Girshick, R.: Mask r-cnn. In: Proceedings
  of the IEEE international conference on computer vision. pp. 2961--2969
  (2017)

\bibitem{hernandez2019human}
Hernandez, A., Gall, J., Moreno-Noguer, F.: Human motion prediction via
  spatio-temporal inpainting. In: Proceedings of the IEEE/CVF International
  Conference on Computer Vision. pp. 7134--7143 (2019)

\bibitem{huang2017towards}
Huang, Y., Bogo, F., Lassner, C., Kanazawa, A., Gehler, P.V., Romero, J.,
  Akhter, I., Black, M.J.: Towards accurate marker-less human shape and pose
  estimation over time. In: 2017 international conference on 3D vision (3DV).
  pp. 421--430. IEEE (2017)

\bibitem{ionescu2013human3}
Ionescu, C., Papava, D., Olaru, V., Sminchisescu, C.: Human3. 6m: Large scale
  datasets and predictive methods for 3d human sensing in natural environments.
  IEEE transactions on pattern analysis and machine intelligence
  \textbf{36}(7),  1325--1339 (2013)

\bibitem{jiang2017seeing}
Jiang, H., Grauman, K.: Seeing invisible poses: Estimating 3d body pose from
  egocentric video. In: 2017 IEEE Conference on Computer Vision and Pattern
  Recognition (CVPR). pp. 3501--3509. IEEE (2017)

\bibitem{joo2020eft}
Joo, H., Neverova, N., Vedaldi, A.: Exemplar fine-tuning for 3d human pose
  fitting towards in-the-wild 3d human pose estimation (2021)

\bibitem{joo2019towards}
Joo, H., Simon, T., Cikara, M., Sheikh, Y.: Towards social artificial
  intelligence: Nonverbal social signal prediction in a triadic interaction.
  In: Proceedings of the IEEE/CVF Conference on Computer Vision and Pattern
  Recognition. pp. 10873--10883 (2019)

\bibitem{joo2017panoptic}
Joo, H., Simon, T., Li, X., Liu, H., Tan, L., Gui, L., Banerjee, S., Godisart,
  T., Nabbe, B., Matthews, I., et~al.: Panoptic studio: A massively multiview
  system for social interaction capture. IEEE transactions on pattern analysis
  and machine intelligence  \textbf{41}(1),  190--204 (2017)

\bibitem{joo2018total}
Joo, H., Simon, T., Sheikh, Y.: Total capture: A 3d deformation model for
  tracking faces, hands, and bodies. In: Proceedings of the IEEE conference on
  computer vision and pattern recognition. pp. 8320--8329 (2018)

\bibitem{kanazawa2018end}
Kanazawa, A., Black, M.J., Jacobs, D.W., Malik, J.: End-to-end recovery of
  human shape and pose. In: Proceedings of the IEEE Conference on Computer
  Vision and Pattern Recognition. pp. 7122--7131 (2018)

\bibitem{kanazawa2019learning}
Kanazawa, A., Zhang, J.Y., Felsen, P., Malik, J.: Learning 3d human dynamics
  from video. In: Proceedings of the IEEE Conference on Computer Vision and
  Pattern Recognition. pp. 5614--5623 (2019)

\bibitem{kay2017kinetics}
Kay, W., Carreira, J., Simonyan, K., Zhang, B., Hillier, C., Vijayanarasimhan,
  S., Viola, F., Green, T., Back, T., Natsev, P., et~al.: The kinetics human
  action video dataset. arXiv preprint arXiv:1705.06950  (2017)

\bibitem{kazakos2019epic}
Kazakos, E., Nagrani, A., Zisserman, A., Damen, D.: Epic-fusion: Audio-visual
  temporal binding for egocentric action recognition. In: Proceedings of the
  IEEE/CVF International Conference on Computer Vision. pp. 5492--5501 (2019)

\bibitem{kitani2011fast}
Kitani, K.M., Okabe, T., Sato, Y., Sugimoto, A.: Fast unsupervised ego-action
  learning for first-person sports videos. In: CVPR 2011. pp. 3241--3248. IEEE
  (2011)

\bibitem{kocabas2020vibe}
Kocabas, M., Athanasiou, N., Black, M.J.: Vibe: Video inference for human body
  pose and shape estimation. In: Proceedings of the IEEE/CVF Conference on
  Computer Vision and Pattern Recognition. pp. 5253--5263 (2020)

\bibitem{Kocabas_PARE_2021}
Kocabas, M., Huang, C.H.P., Hilliges, O., Black, M.J.: {PARE}: Part attention
  regressor for {3D} human body estimation. In: Proceedings International
  Conference on Computer Vision (ICCV). pp. 11127--11137. IEEE (Oct 2021)

\bibitem{Kocabas_SPEC_2021}
Kocabas, M., Huang, C.H.P., Tesch, J., M\"uller, L., Hilliges, O., Black, M.J.:
  {SPEC}: Seeing people in the wild with an estimated camera. In: Proc.
  International Conference on Computer Vision (ICCV). pp. 11035--11045 (Oct
  2021)

\bibitem{kolotouros2019learning}
Kolotouros, N., Pavlakos, G., Black, M.J., Daniilidis, K.: Learning to
  reconstruct 3d human pose and shape via model-fitting in the loop. In:
  Proceedings of the IEEE International Conference on Computer Vision. pp.
  2252--2261 (2019)

\bibitem{kolotouros2019cmr}
Kolotouros, N., Pavlakos, G., Daniilidis, K.: Convolutional mesh regression for
  single-image human shape reconstruction. In: CVPR (2019)

\bibitem{kolotouros2021prohmr}
Kolotouros, N., Pavlakos, G., Jayaraman, D., Daniilidis, K.: Probabilistic
  modeling for human mesh recovery. In: ICCV (2021)

\bibitem{kwon2021h2o}
Kwon, T., Tekin, B., Stuhmer, J., Bogo, F., Pollefeys, M.: {H2O}: {T}wo hands
  manipulating objects for first person interaction recognition. In:
  International Conference on Computer Vision (ICCV) (2021)

\bibitem{cmu}
Lab, C.G.: {CMU Graphics Lab Motion Capture Database}.
  \url{http://mocap.cs.cmu.edu/} (2000)

\bibitem{lee2012discovering}
Lee, Y.J., Ghosh, J., Grauman, K.: Discovering important people and objects for
  egocentric video summarization. In: 2012 IEEE conference on computer vision
  and pattern recognition. pp. 1346--1353. IEEE (2012)

\bibitem{li2019deep}
Li, H., Cai, Y., Zheng, W.S.: Deep dual relation modeling for egocentric
  interaction recognition. In: Proceedings of the IEEE/CVF Conference on
  Computer Vision and Pattern Recognition. pp. 7932--7941 (2019)

\bibitem{li2021hybrik}
Li, J., Xu, C., Chen, Z., Bian, S., Yang, L., Lu, C.: Hybrik: A hybrid
  analytical-neural inverse kinematics solution for 3d human pose and shape
  estimation. In: Proceedings of the IEEE/CVF Conference on Computer Vision and
  Pattern Recognition. pp. 3383--3393 (2021)

\bibitem{li2018eye}
Li, Y., Liu, M., Rehg, J.M.: In the eye of beholder: Joint learning of gaze and
  actions in first person video. In: Proceedings of the European Conference on
  Computer Vision (ECCV). pp. 619--635 (2018)

\bibitem{lin2021end-to-end}
Lin, K., Wang, L., Liu, Z.: End-to-end human pose and mesh reconstruction with
  transformers. In: CVPR (2021)

\bibitem{liu20204d}
Liu, M., Yang, D., Zhang, Y., Cui, Z., Rehg, J.M., Tang, S.: {4D} human body
  capture from egocentric video via {3D} scene grounding. 2021 international
  conference on 3D vision (3DV)  (2021)

\bibitem{loper2015smpl}
Loper, M., Mahmood, N., Romero, J., Pons-Moll, G., Black, M.J.: Smpl: A skinned
  multi-person linear model. ACM transactions on graphics (TOG)
  \textbf{34}(6),  1--16 (2015)

\bibitem{luo20203d}
Luo, Z., Golestaneh, S.A., Kitani, K.M.: 3d human motion estimation via motion
  compression and refinement. In: Proceedings of the Asian Conference on
  Computer Vision (2020)

\bibitem{luo2020kinematics}
Luo, Z., Hachiuma, R., Yuan, Y., Iwase, S., Kitani, K.M.: Kinematics-guided
  reinforcement learning for object-aware 3d ego-pose estimation. arXiv
  preprint arXiv:2011.04837  (2020)

\bibitem{mahmood2019amass}
Mahmood, N., Ghorbani, N., Troje, N.F., Pons-Moll, G., Black, M.J.: Amass:
  Archive of motion capture as surface shapes. In: Proceedings of the IEEE/CVF
  International Conference on Computer Vision. pp. 5442--5451 (2019)

\bibitem{von2018recovering}
von Marcard, T., Henschel, R., Black, M.J., Rosenhahn, B., Pons-Moll, G.:
  Recovering accurate 3d human pose in the wild using imus and a moving camera.
  In: Proceedings of the European Conference on Computer Vision (ECCV). pp.
  601--617 (2018)

\bibitem{vonPon2016a}
von Marcard, T., Pons-Moll, G., Rosenhahn, B.: Human pose estimation from video
  and imus. Transactions on Pattern Analysis and Machine Intelligence
  \textbf{38}(8),  1533--1547 (Jan 2016)

\bibitem{mehta2017monocular}
Mehta, D., Rhodin, H., Casas, D., Fua, P., Sotnychenko, O., Xu, W., Theobalt,
  C.: Monocular 3d human pose estimation in the wild using improved cnn
  supervision. In: 2017 international conference on 3D vision (3DV). pp.
  506--516. IEEE (2017)

\bibitem{Moon_2020_ECCV_I2L-MeshNet}
Moon, G., Lee, K.M.: I2l-meshnet: Image-to-lixel prediction network for
  accurate 3d human pose and mesh estimation from a single rgb image. In:
  European Conference on Computer Vision (ECCV) (2020)

\bibitem{narayan2014action}
Narayan, S., Kankanhalli, M.S., Ramakrishnan, K.R.: Action and interaction
  recognition in first-person videos. In: Proceedings of the IEEE Conference on
  Computer Vision and Pattern Recognition Workshops. pp. 512--518 (2014)

\bibitem{ng2020you2me}
Ng, E., Xiang, D., Joo, H., Grauman, K.: You2me: Inferring body pose in
  egocentric video via first and second person interactions. In: Proceedings of
  the IEEE/CVF Conference on Computer Vision and Pattern Recognition. pp.
  9890--9900 (2020)

\bibitem{northcutt2020egocom}
Northcutt, C., Zha, S., Lovegrove, S., Newcombe, R.: Egocom: A multi-person
  multi-modal egocentric communications dataset. IEEE Transactions on Pattern
  Analysis and Machine Intelligence  (2020)

\bibitem{ogaki2012coupling}
Ogaki, K., Kitani, K.M., Sugano, Y., Sato, Y.: Coupling eye-motion and
  ego-motion features for first-person activity recognition. In: 2012 IEEE
  Computer Society Conference on Computer Vision and Pattern Recognition
  Workshops. pp.~1--7. IEEE (2012)

\bibitem{omran2018neural}
Omran, M., Lassner, C., Pons-Moll, G., Gehler, P., Schiele, B.: Neural body
  fitting: Unifying deep learning and model based human pose and shape
  estimation. In: 2018 international conference on 3D vision (3DV). pp.
  484--494. IEEE (2018)

\bibitem{Patel:CVPR:2021}
Patel, P., Huang, C.H.P., Tesch, J., Hoffmann, D.T., Tripathi, S., Black, M.J.:
  {AGORA}: Avatars in geography optimized for regression analysis. In:
  Proceedings IEEE/CVF Conf.~on Computer Vision and Pattern Recognition
  ({CVPR}) (Jun 2021)

\bibitem{pavlakos2019expressive}
Pavlakos, G., Choutas, V., Ghorbani, N., Bolkart, T., Osman, A.A., Tzionas, D.,
  Black, M.J.: Expressive body capture: 3d hands, face, and body from a single
  image. In: Proceedings of the IEEE Conference on Computer Vision and Pattern
  Recognition. pp. 10975--10985 (2019)

\bibitem{pech2000diatom}
Pech-Pacheco, J.L., Crist{\'o}bal, G., Chamorro-Martinez, J.,
  Fern{\'a}ndez-Valdivia, J.: Diatom autofocusing in brightfield microscopy: a
  comparative study. In: Proceedings 15th International Conference on Pattern
  Recognition. ICPR-2000. vol.~3, pp. 314--317. IEEE (2000)

\bibitem{pirsiavash2012detecting}
Pirsiavash, H., Ramanan, D.: Detecting activities of daily living in
  first-person camera views. In: 2012 IEEE conference on computer vision and
  pattern recognition. pp. 2847--2854. IEEE (2012)

\bibitem{rong2021frankmocap}
Rong, Y., Shiratori, T., Joo, H.: Frankmocap: A monocular 3d whole-body pose
  estimation system via regression and integration. In: IEEE International
  Conference on Computer Vision Workshops (2021)

\bibitem{ryoo2013first}
Ryoo, M.S., Matthies, L.: First-person activity recognition: What are they
  doing to me? In: Proceedings of the IEEE conference on computer vision and
  pattern recognition. pp. 2730--2737 (2013)

\bibitem{saini2019markerless}
Saini, N., Price, E., Tallamraju, R., Enficiaud, R., Ludwig, R., Martinovic,
  I., Ahmad, A., Black, M.J.: Markerless outdoor human motion capture using
  multiple autonomous micro aerial vehicles. In: Proceedings of the IEEE/CVF
  International Conference on Computer Vision. pp. 823--832 (2019)

\bibitem{shiratori2011motion}
Shiratori, T., Park, H.S., Sigal, L., Sheikh, Y., Hodgins, J.K.: Motion capture
  from body-mounted cameras. In: ACM SIGGRAPH 2011 papers, pp. 1--10 (2011)

\bibitem{sigurdsson2018actor}
Sigurdsson, G.A., Gupta, A., Schmid, C., Farhadi, A., Alahari, K.: Actor and
  observer: Joint modeling of first and third-person videos. In: Proceedings of
  the IEEE Conference on Computer Vision and Pattern Recognition. pp.
  7396--7404 (2018)

\bibitem{song2020lgd}
Song, J., Chen, X., Hilliges, O.: Human body model fitting by learned gradient
  descent  (2020)

\bibitem{sun2019human}
Sun, Y., Ye, Y., Liu, W., Gao, W., Fu, Y., Mei, T.: Human mesh recovery from
  monocular images via a skeleton-disentangled representation. In: Proceedings
  of the IEEE International Conference on Computer Vision. pp. 5349--5358
  (2019)

\bibitem{tan2017indirect}
Tan, J.K.V., Budvytis, I., Cipolla, R.: Indirect deep structured learning for
  3d human body shape and pose prediction  (2017)

\bibitem{tome2020selfpose}
Tome, D., Alldieck, T., Peluse, P., Pons-Moll, G., Agapito, L., Badino, H.,
  De~la Torre, F.: Selfpose: 3d egocentric pose estimation from a headset
  mounted camera. arXiv preprint arXiv:2011.01519  (2020)

\bibitem{tome2019xr}
Tome, D., Peluse, P., Agapito, L., Badino, H.: xr-egopose: Egocentric 3d human
  pose from an hmd camera. In: Proceedings of the IEEE/CVF International
  Conference on Computer Vision. pp. 7728--7738 (2019)

\bibitem{Trumble:BMVC:2017}
Trumble, M., Gilbert, A., Malleson, C., Hilton, A., Collomosse, J.: Total
  capture: 3d human pose estimation fusing video and inertial sensors. In: 2017
  British Machine Vision Conference (BMVC) (2017)

\bibitem{tung2017self}
Tung, H.Y., Tung, H.W., Yumer, E., Fragkiadaki, K.: Self-supervised learning of
  motion capture. In: Advances in Neural Information Processing Systems. pp.
  5236--5246 (2017)

\bibitem{hl2_rm}
Ungureanu, D., Bogo, F., Galliani, S., Sama, P., Duan, X., Meekhof, C.,
  Stuhmer, J., Cashman, T.J., Tekin, B., Schonberger, J.L., Tekin, B., Olszta,
  P., Pollefeys, M.: {HoloLens 2 Research Mode as a Tool for Computer Vision
  Research}. arXiv:2008.11239  (2020)

\bibitem{wandt2021canonpose}
Wandt, B., Rudolph, M., Zell, P., Rhodin, H., Rosenhahn, B.: Canonpose:
  Self-supervised monocular 3d human pose estimation in the wild. In:
  Proceedings of the IEEE/CVF Conference on Computer Vision and Pattern
  Recognition. pp. 13294--13304 (2021)

\bibitem{wang2017outdoor}
Wang, Y., Liu, Y., Tong, X., Dai, Q., Tan, P.: Outdoor markerless motion
  capture with sparse handheld video cameras. IEEE transactions on
  visualization and computer graphics  \textbf{24}(5),  1856--1866 (2017)

\bibitem{weng2021holistic}
Weng, Z., Yeung, S.: Holistic 3d human and scene mesh estimation from single
  view images. In: Proceedings of the IEEE/CVF Conference on Computer Vision
  and Pattern Recognition. pp. 334--343 (2021)

\bibitem{xiang2019monocular}
Xiang, D., Joo, H., Sheikh, Y.: Monocular total capture: Posing face, body, and
  hands in the wild. In: Proceedings of the IEEE/CVF Conference on Computer
  Vision and Pattern Recognition (2019)

\bibitem{xu2019mo}
Xu, W., Chatterjee, A., Zollhoefer, M., Rhodin, H., Fua, P., Seidel, H.P.,
  Theobalt, C.: {Mo2Cap2}: {R}eal-time mobile 3{D} motion capture with a
  cap-mounted fisheye camera. IEEE transactions on visualization and computer
  graphics  \textbf{25}(5),  2093--2101 (2019)

\bibitem{xu2019denserac}
Xu, Y., Zhu, S.C., Tung, T.: Denserac: Joint 3d pose and shape estimation by
  dense render-and-compare. In: Proceedings of the IEEE/CVF International
  Conference on Computer Vision. pp. 7760--7770 (2019)

\bibitem{yang2016wearable}
Yang, J.A., Lee, C.H., Yang, S.W., Somayazulu, V.S., Chen, Y.K., Chien, S.Y.:
  Wearable social camera: Egocentric video summarization for social
  interaction. In: 2016 IEEE International Conference on Multimedia \& Expo
  Workshops (ICMEW). pp.~1--6. IEEE (2016)

\bibitem{yonetani2016recognizing}
Yonetani, R., Kitani, K.M., Sato, Y.: Recognizing micro-actions and reactions
  from paired egocentric videos. In: Proceedings of the IEEE Conference on
  Computer Vision and Pattern Recognition. pp. 2629--2638 (2016)

\bibitem{yu2020humbi}
Yu, Z., Yoon, J.S., Lee, I.K., Venkatesh, P., Park, J., Yu, J., Park, H.S.:
  Humbi: A large multiview dataset of human body expressions. In: Proceedings
  of the IEEE/CVF Conference on Computer Vision and Pattern Recognition. pp.
  2990--3000 (2020)

\bibitem{yuan2019ego}
Yuan, Y., Kitani, K.: Ego-pose estimation and forecasting as real-time pd
  control. In: Proceedings of the IEEE/CVF International Conference on Computer
  Vision. pp. 10082--10092 (2019)

\bibitem{yuan2021simpoe}
Yuan, Y., Wei, S.E., Simon, T., Kitani, K., Saragih, J.: Simpoe: Simulated
  character control for 3d human pose estimation. In: Proceedings of the
  IEEE/CVF Conference on Computer Vision and Pattern Recognition. pp.
  7159--7169 (2021)

\bibitem{zanfir2020weakly}
Zanfir, A., Bazavan, E.G., Xu, H., Freeman, B., Sukthankar, R., Sminchisescu,
  C.: Weakly supervised 3d human pose and shape reconstruction with normalizing
  flows. arXiv preprint arXiv:2003.10350  (2020)

\bibitem{zhang2021body}
Zhang, J., Yu, D., Liew, J.H., Nie, X., Feng, J.: Body meshes as points. In:
  Proceedings of the IEEE/CVF Conference on Computer Vision and Pattern
  Recognition. pp. 546--556 (2021)

\bibitem{Zhang:ICCV:2021}
Zhang, S., Zhang, Y., Bogo, F., Marc, P., Tang, S.: Learning motion priors for
  4d human body capture in 3d scenes. In: International Conference on Computer
  Vision (ICCV) (Oct 2021)

\bibitem{zhang20204d}
Zhang, Y., An, L., Yu, T., Li, X., Li, K., Liu, Y.: 4d association graph for
  realtime multi-person motion capture using multiple video cameras. In:
  Proceedings of the IEEE/CVF Conference on Computer Vision and Pattern
  Recognition. pp. 1324--1333 (2020)

\bibitem{zhang2022can}
Zhang, Z., Crandall, D., Proulx, M., Talathi, S., Sharma, A.: Can gaze inform
  egocentric action recognition? In: 2022 Symposium on Eye Tracking Research
  and Applications. pp.~1--7 (2022)

\bibitem{zhou2021monocular}
Zhou, Y., Habermann, M., Habibie, I., Tewari, A., Theobalt, C., Xu, F.:
  Monocular real-time full body capture with inter-part correlations. In:
  Proceedings of the IEEE/CVF Conference on Computer Vision and Pattern
  Recognition. pp. 4811--4822 (2021)

\end{thebibliography}

\clearpage

\appendix

\begin{center}
\large{\bf EgoBody: Human Body Shape and Motion of \\ Interacting People from Head-Mounted Devices \\
\vspace{0.3cm} 
**Supplementary Material**\\
}
\vspace{1cm}
\end{center}

\setcounter{page}{1}
\setcounter{table}{0}
\setcounter{figure}{0}
\renewcommand{\thetable}{S\arabic{table}}
\renewcommand\thefigure{S\arabic{figure}}

\section{Details of Dataset Building}
\subsection{Calibration}
To spatially calibrate the Kinects-Kinects and Kinects-HoloLens2, we employ a checkerboard to obtain an initial calibration; we then refine the result by running ICP~\cite{icp} on the scene point clouds reconstructed from the depth sensor of the devices. 
Additionally, the Kinects-HoloLens2 calibration is refined by the keypoint-based optimization scheme as described in the main paper Sec.~3.3.
To register camera data into the 3D scene, we first manually annotate a set of correspondence points between the scene mesh and scene point clouds given by the Kinect depth frames to obtain an initial rigid transformation, which is again refined via ICP.

\subsection{SMPL-X Body Model}
We use SMPL-X~\cite{pavlakos2019expressive}, which represents the body as a function $\mathcal{M}_b(\boldsymbol{\gamma}, \boldsymbol{\beta}, \boldsymbol{\theta}, \boldsymbol{\phi})$. It maps global translation $\boldsymbol{\gamma} \in \mathbb{R}^3$, body shape $\boldsymbol{\beta} \in \mathbb{R}^{10}$, pose $\boldsymbol{\theta}$ and facial expression $\boldsymbol{\phi} \in \mathbb{R}^{10}$ to a triangle body mesh with 10,475 body vertices.
$\mathcal{M}_b=(V_b, F_b)$ with body vertices $V_b \in \mathbb{R}^{10475\times 3}$ and faces $F_b$.
The pose parameters include body, facial and hand poses. 
$J(\boldsymbol{\beta})$ denotes the 3D body joints in the neutral pose, which can then be posed according to a given $\boldsymbol{\theta}$. 

\subsection{Data Processing}
\myparagraph{Body point cloud extraction.} We use Mask-RCNN~\cite{he2017mask} to get coarse human instance segmentation masks in Kinect RGB frames and refine the masks with DeepLabv3~\cite{he2017mask}. The obtained human instance segmentation masks are mapped to Kinect depth frames to segment the human body point clouds from point clouds extracted from depth. 

\myparagraph{Subject index reordering.}
Note that OpenPose~\cite{openpose}, DeepLabv3 and Mask-RCNN all work per-frame, without temporal tracking. For each sequence, we therefore reorder subject indices from OpenPose detections and human masks by their relative position to each other in 2D, such that each subject has a consistent index across all frames and all Kinect views.

\begin{figure}[h]
    \centering
    \includegraphics[width=\linewidth]{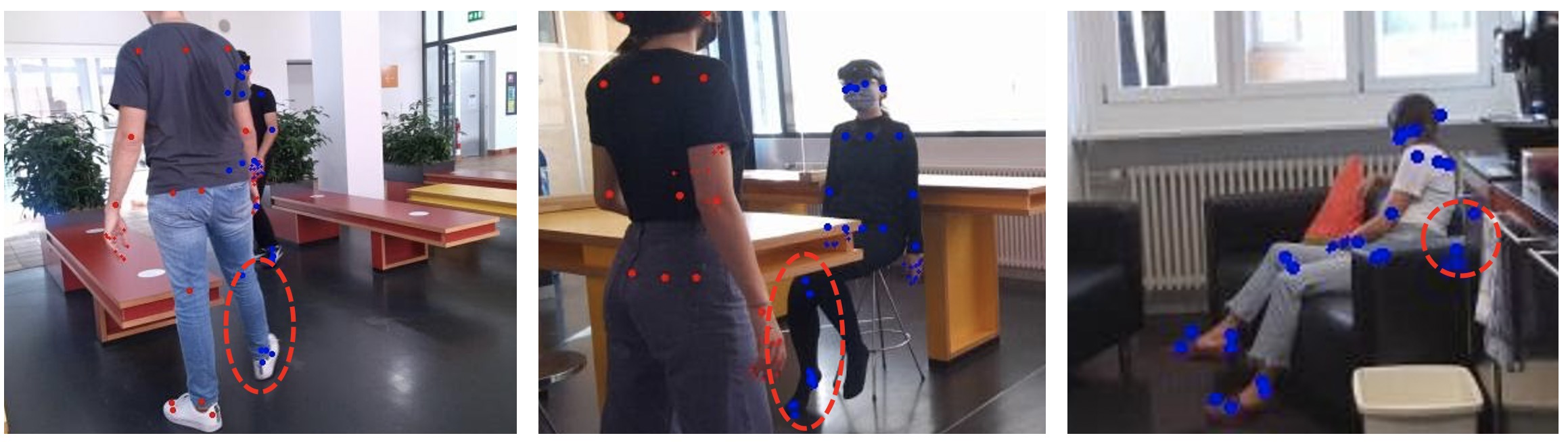}
    \caption{Limitations for OpenPose 2D joint detection when body-body occlusion (left), body-scene occlusion (middle), or self-occlusion (right) happens. Different keypoint colors denote different detected body indices.}
    \label{fig:openpose-failure}
\end{figure}

\myparagraph{Data cleaning.}
2D joint detection and human instance segmentation can fail in the presence of body-body or body-scene occlusions (Fig.~\ref{fig:openpose-failure} left/middle). Thus we manually clean the failed detections and exclude them from the reconstruction pipeline. We also manually clean inaccurate 2D joint detections due to self-occlusions (Fig.~\ref{fig:openpose-failure} right).
We also leverage the depth information from Kinect cameras, to filter out 2D joints with a large difference between its depth value and the median depth value of all 2D joints of the target person.

\myparagraph{EgoSet-\interactee{} subset selecting.}
Egocentric image frames with extreme human body truncations are excluded from EgoSet-\interactee{} subset by the following filtering procedure.
We run OpenPose 2D joint detection on all egocentric image frames, and manually exclude spurious detections (irrelevant people in the background or false positives on scene objects) for each frame. 
As OpenPose may split the joints from the same body into several detections, we merge them into one body in such case, where the joint conflict is resolved by the confidence score. 
We include the frames in EgoSet-\interactee{} subset when at least six valid joints (OpenPose BODY\_25 format) of the \interactee{} are detected. We threshold the joint confidence score by 0.2 as in~\cite{kolotouros2019learning}. Note that five joints concentrated on the head are considered as one joint due to the close distances, as well as the four joints on each foot. The bounding box is computed with the processed results.

\section{Ground-truth Annotation Quality}

\subsection{Reconstruction Accuracy} 
To evaluate the 2D accuracy of the reconstructed body in the egocentric view, we randomly select 1,517 frames and manually annotate their 2D joints via Amazon Mechanical Turk (AMT), using the SMPL-X skeleton definition. 
By projecting 3D joints estimated by our pipeline on the egocentric view, the mean 2D joint error (2D Euclidean distance between the projections and AMT annotations) is 31.08 pixels (image in $1920\times1080$ resolution).
We also evaluate the 3D accuracy of our ground truth on two metrics:
\textbf{Scan-to-body Chamfer Distance} (\textit{1.65cm}), and \textbf{3D per joint error (PJE)} (\textit{4.32cm}). 
The 3D PJE is computed as follows:
we leverage AMT to annotate 2D body joints in all Kinect views for 100 frames, from which the annotated 3D joints are obtained via multi-view triangulation. 
The 3D PJE is then measured between our ground truth SMPL-X body joints and the annotated 3D joints.

\subsection{Motion Smoothness}
The motion smoothness of our dataset is evaluated with the Power Spectrum KL divergence score (PSKL)~\cite{hernandez2019human} as in~\cite{Zhang:ICCV:2021}: we measure the distance between the distribution of joint accelerations in our dataset and that in the high-quality mocap dataset AMASS~\cite{mahmood2019amass}. The lower the score is, the more the motions resemble the natural motions in AMASS. 
Besides, we compare with HPS~\cite{guzov2021human}, a recent egocentric view dataset (Tab.~\ref{tab:eval-smoothness}). The significantly lower PSKL score of \name{} reflects the high quality of our ground truth motions.

\begin{table}[t]
\centering
\caption{Motion smoothness evaluation of HPS and our dataset. PSKL(X,~A) denotes PSKL(HPS/ours,~AMASS), and PSKL(A,~HPS/ours) the reverse direction. Better result in boldface.}
\begin{tabular}{lcc}
\toprule[1pt]
    & PSKL(X,A) $\downarrow$ & PSKL(A,X) $\downarrow$ \\
\midrule
HPS~\cite{guzov2021human}  & 0.924 & 1.044 \\
EgoBody & \textbf{0.312} & \textbf{0.262} \\  
\bottomrule[1pt] 
\end{tabular}
\label{tab:eval-smoothness}
\end{table}

\section{More Statistics}

\begin{figure*}[t]
\centering
    \begin{minipage}[b]{0.49\linewidth}
    \centering  
    \subfloat[]{
    \includegraphics[width=\linewidth]{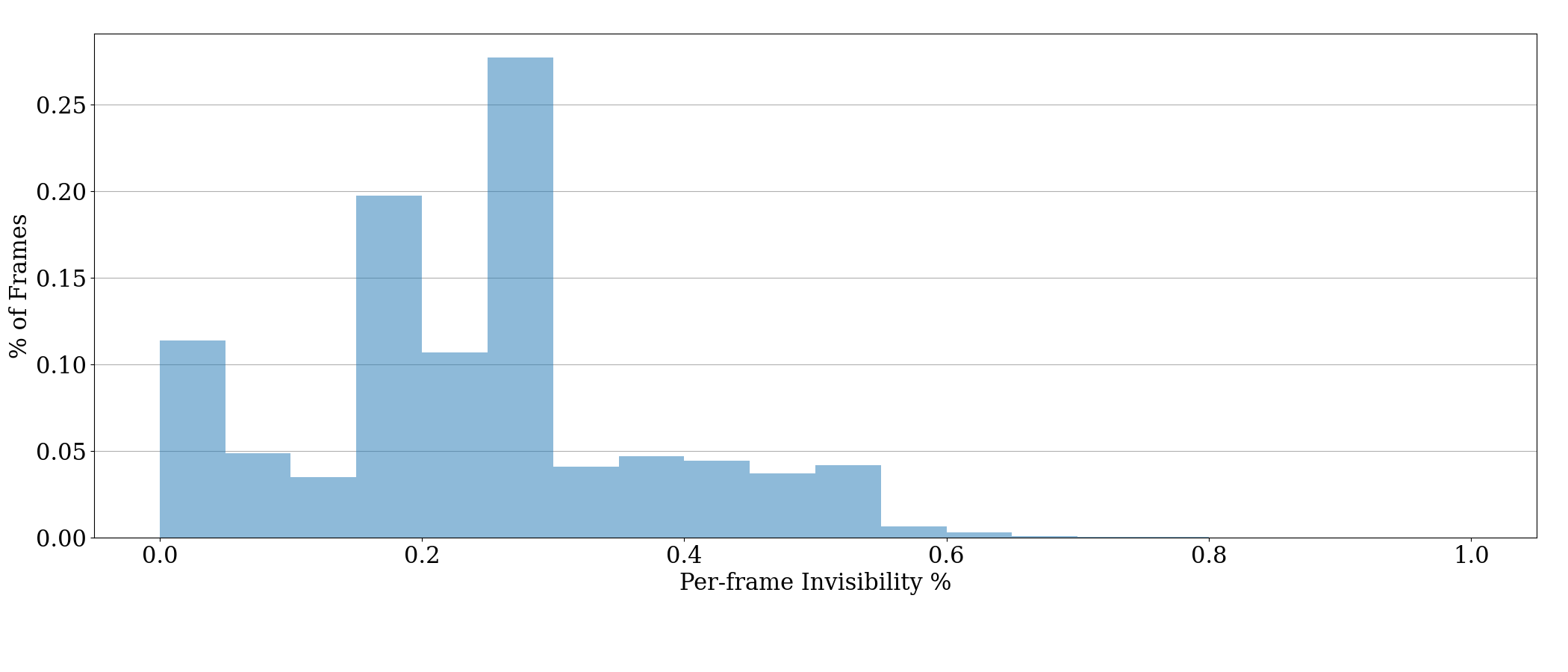}
    }
    \end{minipage}
    \begin{minipage}[b]{0.49\linewidth}
    \centering  
     \subfloat[]{
    \includegraphics[width=\linewidth]{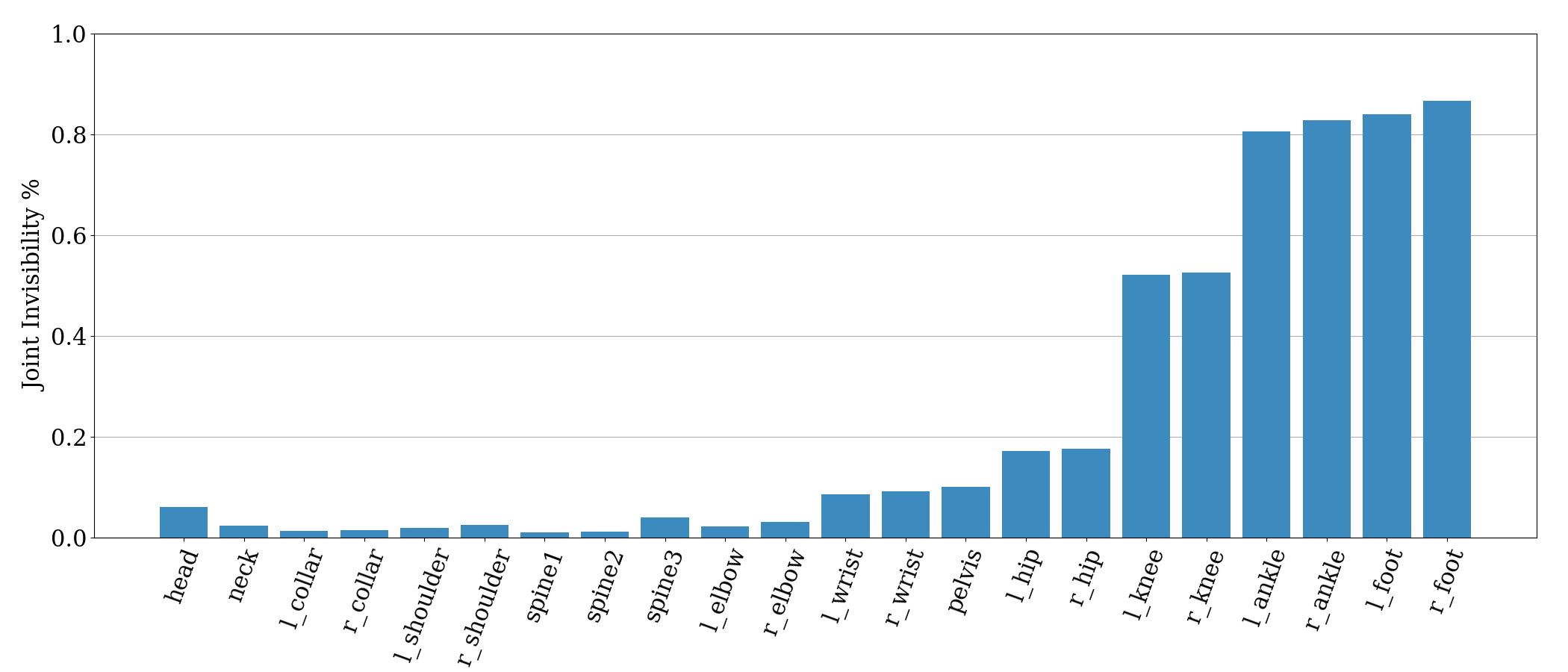}
    }
    \end{minipage}
\caption{(a) Distribution of frame-wise invisibility ratio (\% of invisible joints among all body joints for each frame). (b) Joint-wise invisibility ratio (\% of occurrences when the corresponding joint is invisible among all frames.): `l\_' denotes `left\_' and `r\_' denotes `right\_'.}
\label{fig:joint-vis-stats}
\end{figure*}

\myparagraph{Joint invisibility.}
We consider 2 types of ``invisibility'' measurements: frame-wise invisibility and joint-wise invisibility ratio. 
For each frame, the frame-wise invisibility ratio calculates the percentage of invisible joints among all body joints. As shown in Fig.~\ref{fig:joint-vis-stats}(a), partial body invisibility occurs in most frames, with even over 60\% body joints invisible in extreme cases. 
For each body joint, the joint-wise invisibility ratio calculates the ratio of frames when the joint is invisible among all frames (See Fig.~\ref{fig:joint-vis-stats}(b)). 
The lower part of the body exhibits higher chances of invisibility (knees around 50\% and feet around 80\%). The upper body parts are more visible: neck, shoulder, spine, and elbow joints above all, while wrist and head joints have slightly higher invisibility (around 10\%). 

\begin{figure*}[t]
\centering
    \begin{minipage}[b]{0.49\linewidth}
    \centering  
    \subfloat[]{
    \includegraphics[width=\linewidth]{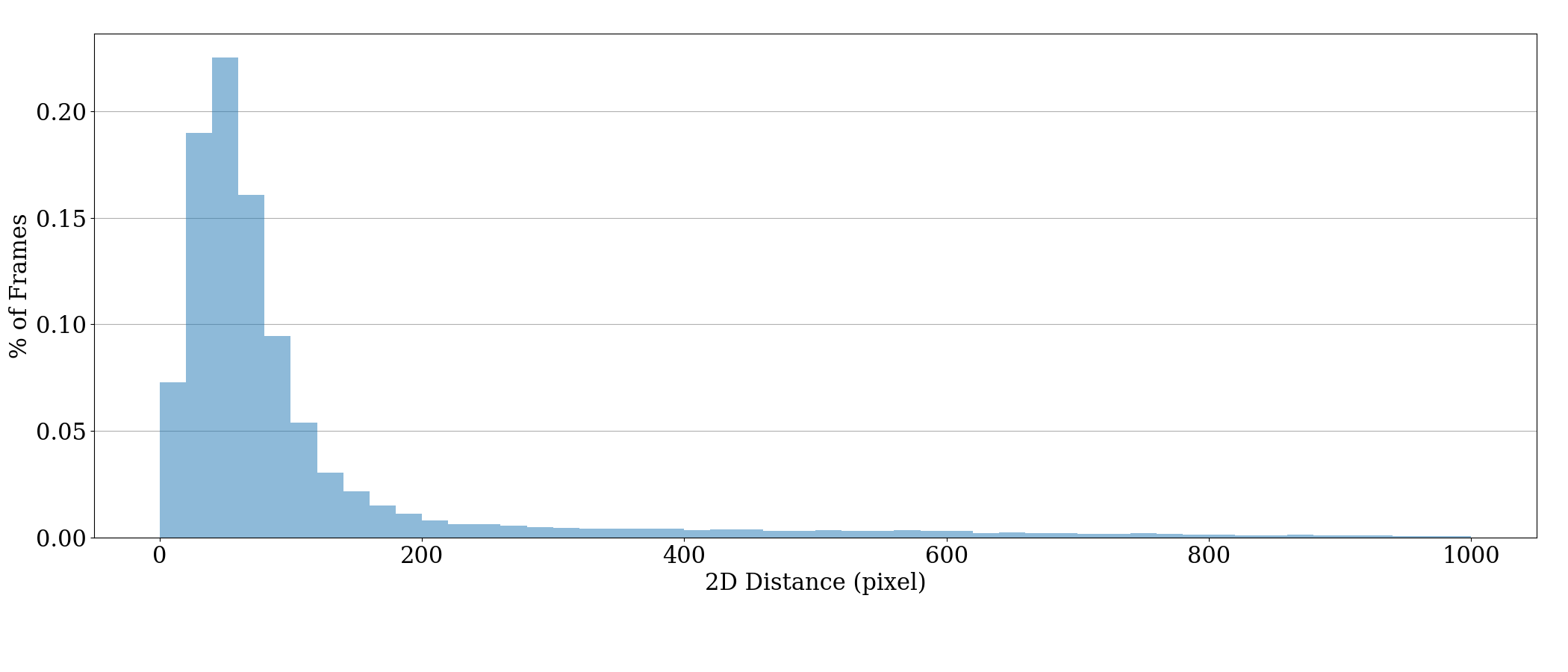}
    }
    \end{minipage}
    \begin{minipage}[b]{0.49\linewidth}
    \centering  
    \subfloat[]{
    \includegraphics[width=\linewidth]{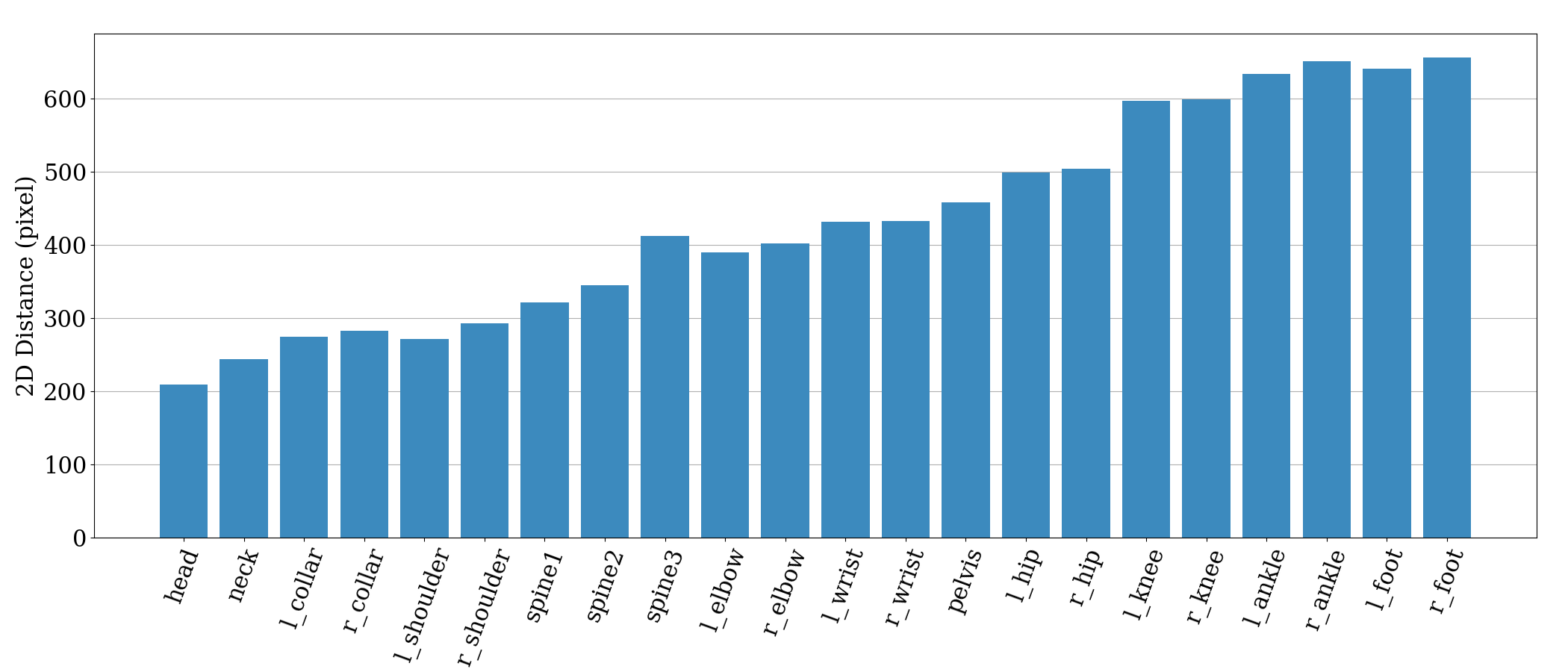}
    }
    \end{minipage}
\caption{(a) Distribution of the 2D distance between the 2D gaze point and its nearest body joint. (b) Mean 2D distance between the 2D gaze point and each body joint. `l\_' denotes `left\_' and `r\_' denotes `right\_'.}
\vspace{-10pt}
\label{fig:gaze-stats}
\end{figure*}

\myparagraph{Eye gaze and attention.}
The HoloLens2 eye tracking provides the eye gaze 3D ray's starting point and orientation, which we can intersect with our 3D reconstructions to calculate the location the user looks at. By projecting the 3D eye gaze point onto the egocentric images,
we perform analysis on the distances between this projected 2D eye gaze point (attention area) and the body joints of the \interactee{} on the egocentric view images.
For all frames where the 2D gaze point lies within the image,
Fig.~\ref{fig:gaze-stats}(a) plots the distribution of the Euclidean distance between the 2D gaze point and its nearest body joint. For more than $75\%$ of the frames, the distance between the 2D gaze point and its nearest joint lies within 120 (pixels), indicating that the camera wearer's attention is highly focused on the \interactee{} during interactions.
The mean distance between each joint and the 2D gaze point over all frames (Fig.~\ref{fig:gaze-stats}(b)) reveals that the subjects' attention tends to be closer to the upper body joints during interactions.

\begin{figure*}[t]
\centering
    \begin{minipage}[b]{0.49\linewidth}
    \centering  
     \subfloat[]{
    \includegraphics[width=\linewidth]{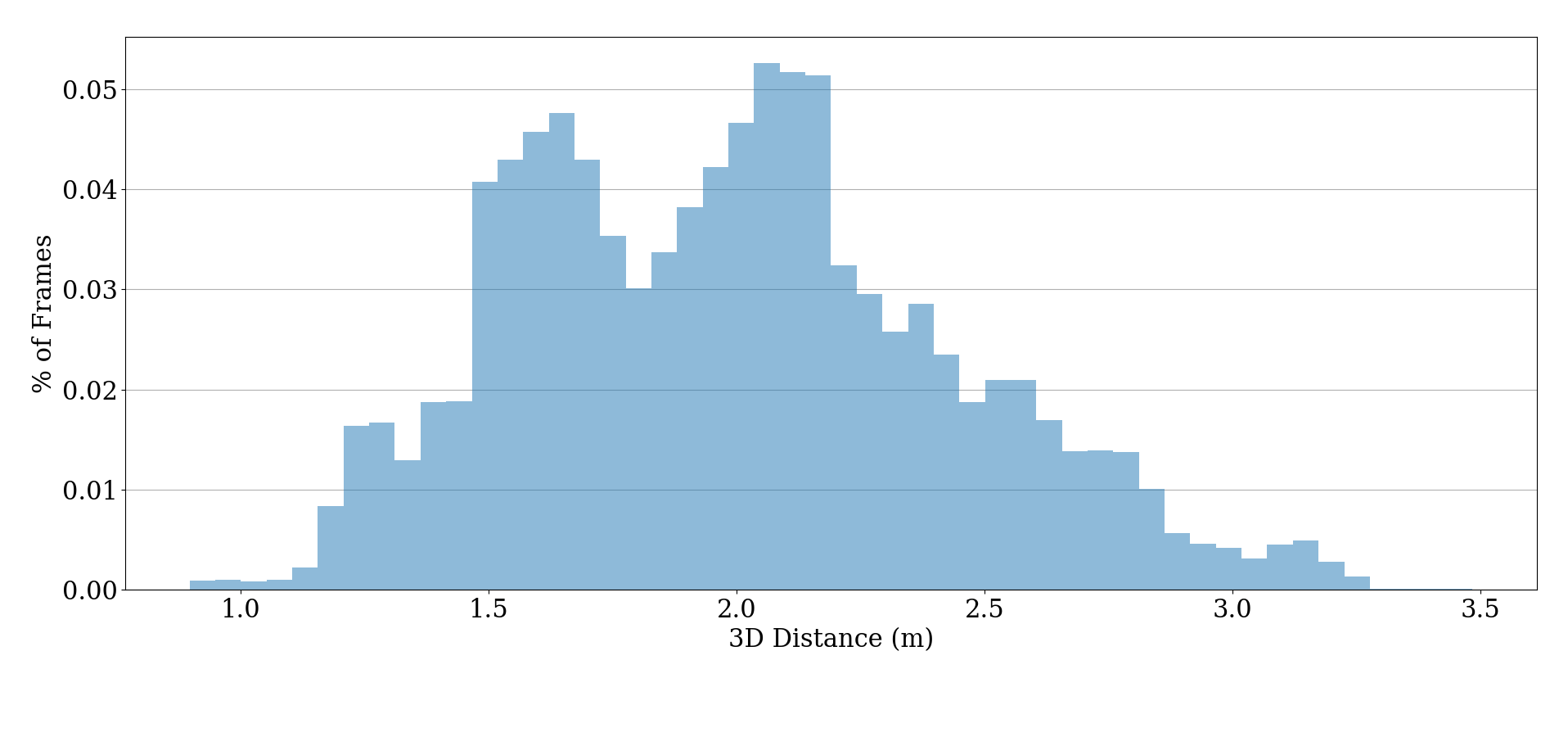}
    }
    \end{minipage}
    \begin{minipage}[b]{0.49\linewidth}
    \centering  
     \subfloat[]{
    \includegraphics[width=\linewidth]{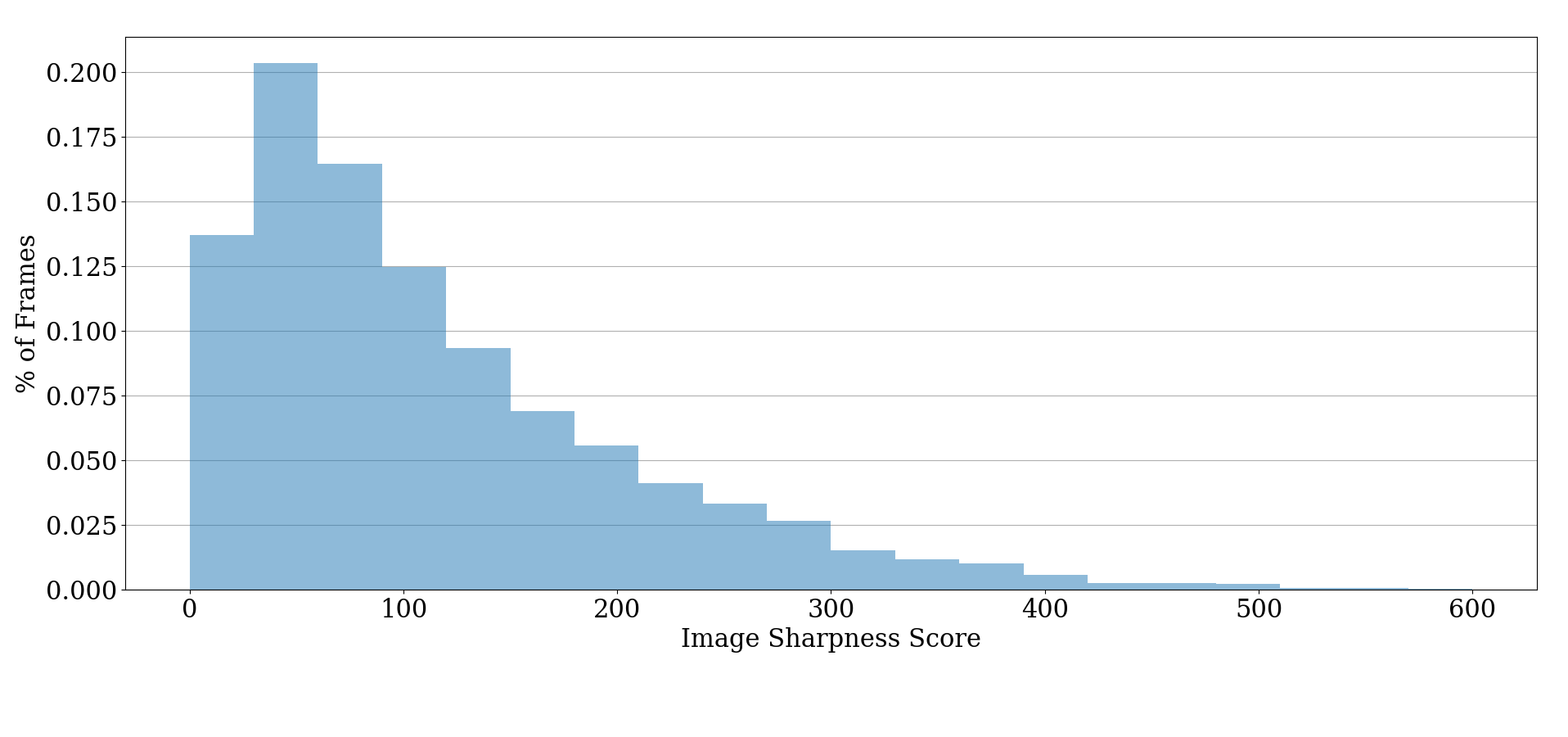}
    }
    \end{minipage}
\caption{(a) Distribution of interaction distances between two interacting subjects. (b) Distribution of the image sharpness score on EgoSet-\interactee{} test set.}
\label{fig:interaction-blur-dist}
\end{figure*}

\myparagraph{Interaction distance.}
\name{} covers a large range of indoor interaction distances: the 3D Euclidean distance between the pelvis joints of the interacting subjects ranges from 0.90\textit{m} to 3.48\textit{m}.
Fig.~\ref{fig:interaction-blur-dist}(a) shows the distribution of the interaction distances between 2 interacting subjects.

\myparagraph{Motion blur.}
The image sharpness score (variance of the Laplacian of the image)~\cite{pech2000diatom} quantifies the motion blur in our dataset (the distribution is shown in Fig.~\ref{fig:interaction-blur-dist}(b)). A higher score indicates sharper images. 

\begin{figure}[ht]
    \centering
    \includegraphics[width=\linewidth]{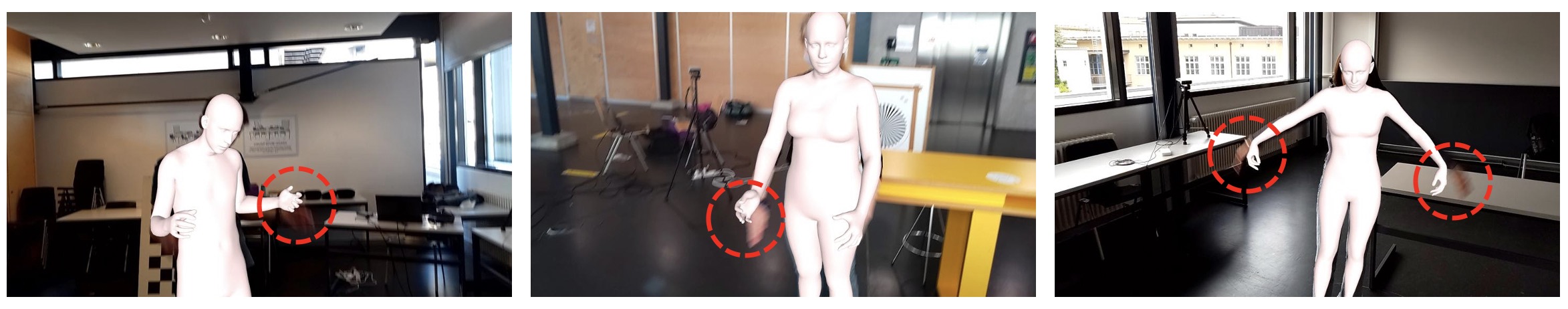}
    \caption{Misalignments between the body mesh and HoloLens2 images during fast motions (for example, fast hand movements).}
    \label{fig:sync-problem}
\end{figure}

\section{Experiments: Details and Discussions}
\subsection{Discussion on Evaluation Metrics} 
While MPJPE is a commonly used measurement for pose estimation accuracy, it is very sparse and does not penalize the error caused by wrong joint twisting.
This motivates us to also include the V2V error as a metric in our benchmark: it is not only a straightforward measure for the shape estimation, but also measures the pose error in a denser and stricter way than MPJPE as it also penalizes erroneous longitudinal joint rotations.

By default we consider the MPJPE and V2V errors without the Procrustes Alignment (PA), as PA eliminates the discrepancy in the global orientation, a major source of errors for most methods. Deprecating PA-based metrics is becoming a recent trend~\cite{Kocabas_PARE_2021,Patel:CVPR:2021}.

\subsection{Baseline Improvement on EgoBody}
\myparagraph{Implementation details.} We fine-tune SPIN~\cite{kolotouros2019learning}, METRO~\cite{lin2021end-to-end}  and EFT~\cite{joo2020eft}   using the official codes, but with slight customization in the training as follows. 
For SPIN, we disable the SMPLify-in-the-loop during training since the \name{} training set already provides direct 3D supervision from the pseudo ground truth. Note that this is in fact the default setting in SPIN when the 3D ground truth is available.

\begin{figure*}[!h]
\vspace{-8pt}
    \centering
    \includegraphics[width=\linewidth]{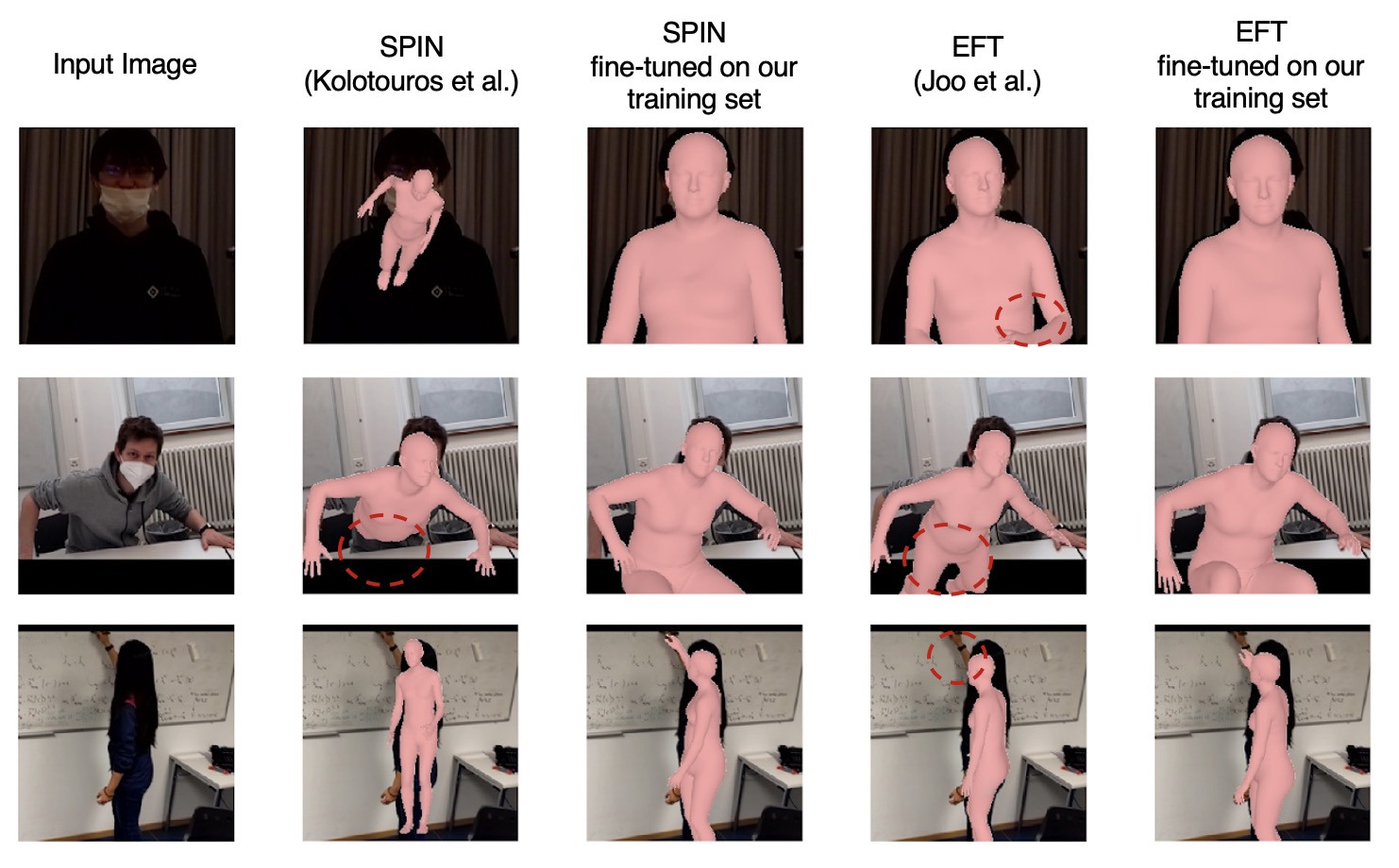}
    \caption{Qualitative results from the baseline evaluation.}
    \label{fig:baseline_qualit}
\end{figure*}

\myparagraph{Extended results and discussions.}
As shown in Fig.~\ref{fig:baseline_qualit}, while SPIN works well on images when the full body is visible, it fails on images where the subjects are truncated. EFT, in contrast, is more robust against such truncation as the model is trained on aggressively cropped images as data augmentation during training.
The effectiveness of EFT's data augmentation is further supported by our fine-tuning experiments.  
After fine tuning SPIN and EFT on our training set, both models show greater robustness against motion blur and image truncation, and quantitatively achieve lower errors than the original models on all metrics, as shown in the main paper Tab.~3.
Together with the experiments on the You2Me dataset (see main paper Sec.~5.4), this shows that our training set can help adapt existing 3DHPS models to egocentric view data.

\subsection{Details of the Cross-dataset Evaluation on You2Me}
Here we report the experimental setup of the You2Me dataset experiment. 
The You2Me dataset~\cite{ng2020you2me} provides egocentric view images (taken with a chest-mounted GoPro camera), and the ground truth 3D joint locations of both the camera wearer and the \interactee{}.
The ground truth 3D joints are however in a world coordinate system, making it infeasible to compute the translation-only MPJPE (see main paper Sec.~5.1): since the camera calibration between the GoPro and the world coordinate is unknown, even a perfect prediction (in the camera coordinate system) may differ from the ground truth up to a rigid transformation. 
To account for this problem, we first perform the Procrustes Alignment, which solves for the scale, translation and global rotation, to align the predicted 3D body joints with the ground truth, and then compute the MPJPE of the aligned bodies, resulting in the PA-MPJPE errors reported in the paper.

\subsection{Experiment with Motion Blur Augmentation}
Data augmentation could potentially simulate blurring and truncation. Can the performance of existing methods be enhanced on EgoBody by simply fine-tuning them on the original dataset that they are trained on with extra data augmentation? 
EFT is trained with aggressive \textit{image cropping} which to an extent simulates body truncation in our dataset. Indeed its superior performance has proven the effect of data augmentation, but such augmentation does not fully address the challenges in EgoBody: a clear performance gap can be seen between the original EFT and all models fine-tuned on our dataset (SPIN-ft, METRO-ft, EFT-ft, see Tab.~4 in main paper). 
Likewise, here we additionally analyze motion blur augmentation.  
We fine-tune the pre-trained SPIN model with additional motion blur augmentation on the datasets it's originally trained on.
For the motion blur we randomly set blur direction, angle, and kernel size to blur the training images with a probability of 0.5 during fine-tuning, and we experiment with multiple settings with different kernel sizes. No improvements are observed compared with the original SPIN model when evaluated on our egocentric test set (Tab.~\ref{tab:spin-blur}). 
Both observations indicate that existing data augmentation techniques cannot fully resolve the challenges in the egocentric setup, and EgoBody fills this gap. 

\begin{table}[!h]
\centering
\caption{\textbf{Evaluation of motion blur augmentation on our egocentric test set}. `SPIN-blur' denotes the result of fine-tuning SPIN on its original training set with additional motion blur augmentation.} 
\begin{tabular}{lcccc}
\toprule[1pt]
 & MPJPE~$\downarrow$ & PA-MPJPE~$\downarrow$ & V2V~$\downarrow$ & PA-V2V~$\downarrow$ \\
\midrule[1pt]
 SPIN~\cite{kolotouros2019learning} & 182.8 & 116.6 & 187.3 & 123.7 \\
 SPIN-blur & 184.9 & 128.2 & 188.5 & 129.2 \\
\bottomrule[1pt]
\end{tabular}
\label{tab:spin-blur}
\end{table}

\subsection{Details for Baseline Methods}
\myparagraph{CMR}~\cite{kolotouros2019cmr} firstly regresses 3D locations of SMPL body vertices via a graph convolutional network. An image-based CNN encodes the input image into a feature vector, which is attached to the graph network defined by a mesh template. A Multi-Layer Perceptron (MLP) predicts the SMPL parameters based on regressed body vertices.

\myparagraph{METRO}~\cite{lin2021end-to-end} adopts the model-free formulation to estimate body vertices and 3D body joints directly. A transformer encoder models interactions for vertex-vertex and vertex-joint via self-attention mechanism.

\myparagraph{SPIN}~\cite{kolotouros2019learning} integrates iterative optimization loops into the neural network training to combine advantages of both regression-based and optimization-based methods. The optimization fits the SMPL body model to 2D joints on the image to enable more robust supervision for the regressor.

\myparagraph{EFT}~\cite{joo2020eft} augments existing large-scale 2D datasets with 3D annotations by Exemplar Fine-Tuning. Starting from a pre-trained 3D pose regressor, the model weights are fine-tuned by fitting 2D joints to images. Supervised by the obtained 3D annotations, a model with the same architecture as SPIN~\cite{kolotouros2019learning} is trained with extreme crop augmentation and auxiliary input representations. 

\myparagraph{LGD}~\cite{song2020lgd} proposes to use neural networks to predict the parameter update rules in the optimization framework. A Gradient Updating Network regresses the update step for SMPL parameters in each optimization iteration.

\myparagraph{PARE}~\cite{Kocabas_PARE_2021} leverages the visibility information of each body part, and predicts body-part-guided attention masks to achieve robust prediction for SMPL parameters with body occlusions.

\section{AMT Annotation Details}

\begin{figure*}[ht]
\centering
    \begin{minipage}[b]{0.58\linewidth}
    \centering  
     \subfloat[]{
    \includegraphics[width=\linewidth]{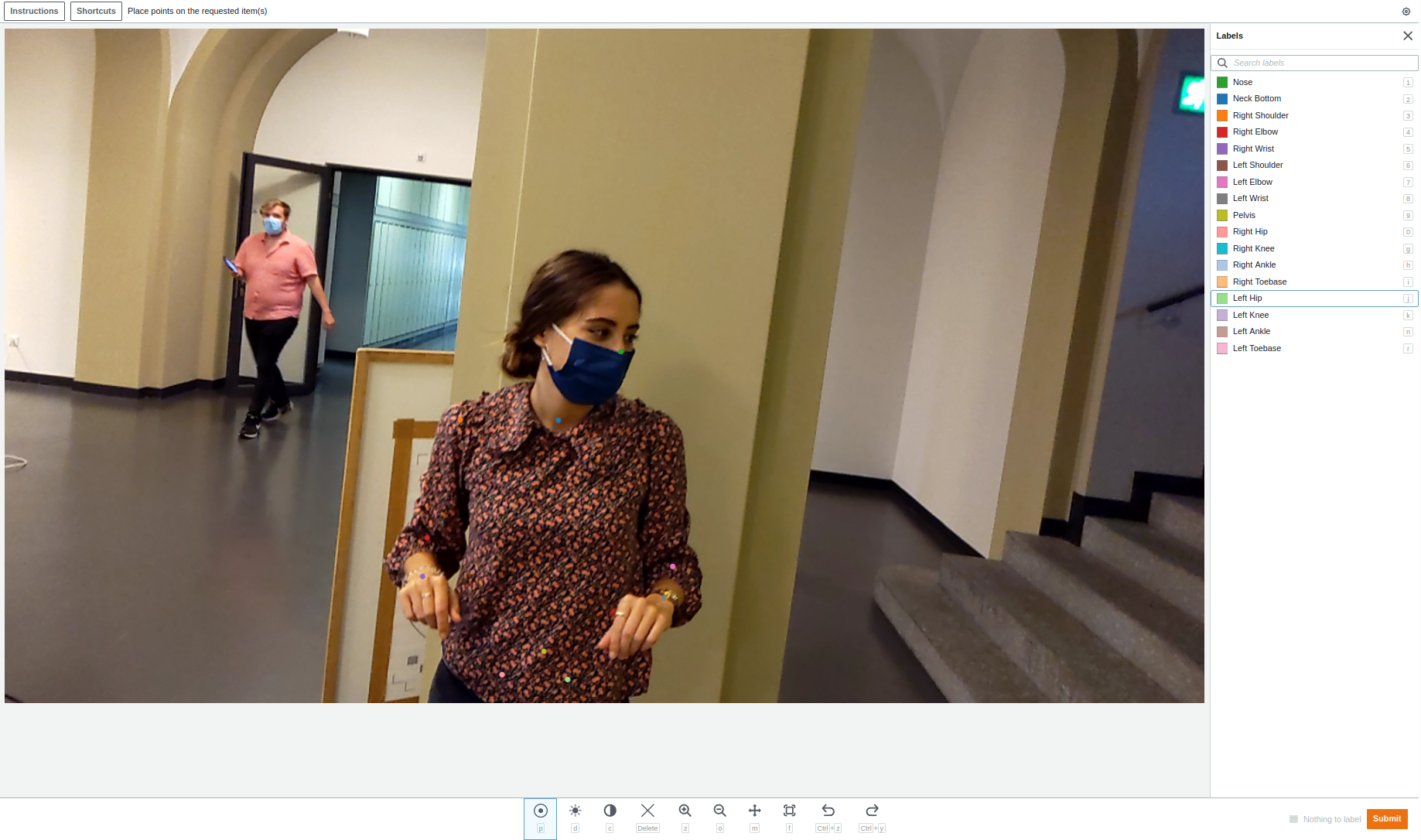} 
    }
    \end{minipage}
    \begin{minipage}[b]{0.38\linewidth}
    \centering  
     \subfloat[]{
    \includegraphics[width=\linewidth]{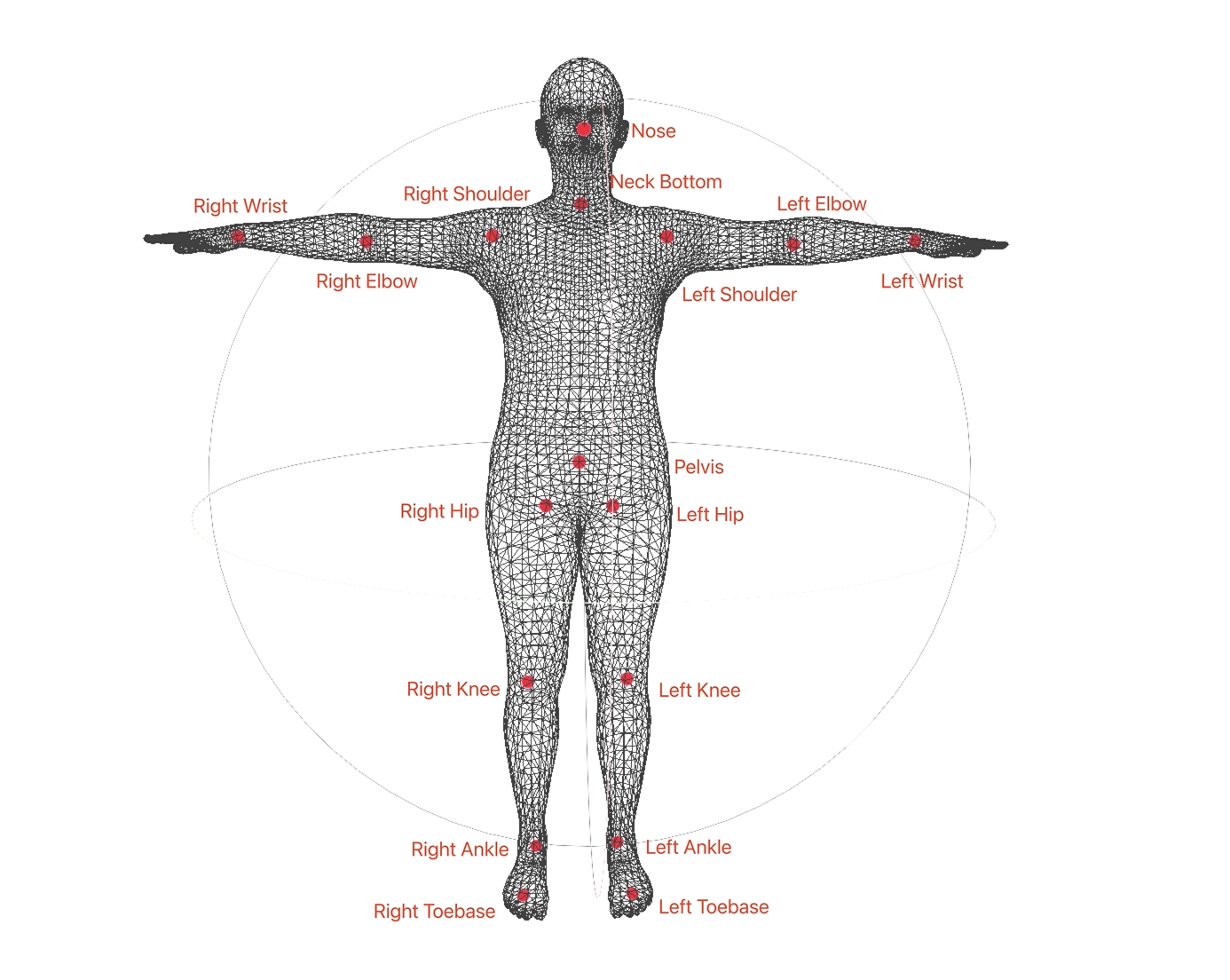}
    }
    \end{minipage}
\caption{(a) The user interface, and the (b) definition of the 17 joints, for AMT manual annotation.}
\label{fig:amt_annotation}
\end{figure*}

To evaluate the body shape and pose annotation accuracy, we collect manual annotations of 2D locations of 17 body joints on the EgoSet-\interactee{} frames via Amazon Mechanical Turk (AMT). The user interface is illustrated in Fig.~\ref{fig:amt_annotation}(a). 
We exclude body joints that are ambiguous for manual annotating (head, spine1, spine2, spine3, left\_collar, right\_collar) from the first 22 SMPL-X body joints, and add the nose joint which is easy to define for users.
The user is provided with the definition of body joints (Fig.~\ref{fig:amt_annotation}(b)), and an image of the target person to annotate (in case of irrelevant people in the background). 
Self-occluded keypoints need to be inferred, while keypoints occluded by scene objects are not required to be annotated.
We downsample with a rate of 50 on the EgoSet-\interactee{} data, which yields a total number of 2,286 frames.

For better annotation quality, each frame is annotated by five users, and joints annotated by less than three users are ignored. A small part of the users flip the left and right side, inducing non-negligible noise for the ground-truth evaluation. To address this issue, we filter out the outliers and correct the flipped annotations by the following procedure. 
For each annotated joint, the 2D distance from each annotation $\mathbf{x}_i$ ($i=0,1,...,4$) to the mean location $\bar{\mathbf{x}}$ is calculated as $d_i=||\mathbf{x}_i-\bar{\mathbf{x}}||_2$. An annotation is considered as the outlier if $\frac{d_i-\bar{d}}{\sigma}>1.5$, where $\bar{d}$ is the average distance, and $\sigma=\sqrt{\frac{1}{n}(d_i-\bar{d})^2}$ is the standard deviation of the distance. For joints that have the counterpart on the other side of body we flip the left/right side of the annotations and perform the same outlier detection, to fix cases when the users flip the left and right side.

\section{Limitations}

As there exists no solution to synchronize HoloLens2 and Kinect via hardware, we align their clocks via software, using a flashlight which is visible to all devices as signal for the first frame. Although HoloLens2 exhibits frame drops occasionally, the corresponding frames of all devices can be aligned according to the timestamps provided by HoloLens2 Research Mode API~\cite{hl2_rm}.
Besides, we empirically observe a small temporal misalignment. In our case, the misalignment can be observed for fast motions (for example, for hand movements, as shown in Fig.~\ref{fig:sync-problem}). However, this issue is inevitable for the synchronization between third-person view cameras and HMDs~\cite{ng2020you2me}.
Despite the small misalignment, our reconstruction reaches a high accuracy as proved by the reconstruction accuracy in Sec.~4.2.

\end{document}